\documentclass[12pt,draftcls,onecolumn]{IEEEtran}
\usepackage{amsfonts}

\usepackage{setspace}
\usepackage{setspace}
\usepackage{amssymb}
\usepackage{epsfig}
\usepackage{graphicx}
\usepackage{xcolor}
\usepackage{algorithm}
\usepackage{algorithmicx}
\usepackage{algpseudocode}
\usepackage{amsmath}
\usepackage{tabularx}
\usepackage{subfig}

\floatname{algorithm}{Algorithm}

\newcommand{\x}{\mathtt{x}}

\begin{document}

\title{A Deep Actor-Critic Reinforcement Learning Framework for Dynamic Multichannel Access}

\author{Chen Zhong, Ziyang Lu, M. Cenk Gursoy, and Senem Velipasalar
\thanks{The authors are with the Department of Electrical
Engineering and Computer Science, Syracuse University, Syracuse, NY, 13244
(e-mail: czhong03@syr.edu, zlu112@syr.edu, mcgursoy@syr.edu, svelipas@syr.edu).}
\thanks{The material in this paper was presented in part at the 2018 IEEE Global Conference on Signal and Information Processing (GlobalSIP), in Nov. 2018.
}}

\date{}

\maketitle

\thispagestyle{empty}

\begin{abstract}
	To make efficient use of limited spectral resources, we in this work propose a deep actor-critic reinforcement learning based framework for dynamic multichannel access. We consider both a single-user case and a scenario in which multiple users attempt to access channels simultaneously. We employ the proposed framework as a single agent in the single-user case, and extend it to a decentralized multi-agent framework in the multi-user scenario.  In both cases, we develop algorithms for the actor-critic deep reinforcement learning and evaluate the proposed learning policies via experiments and numerical results. In the single-user model, in order to evaluate the performance of the proposed channel access policy and the framework's tolerance against uncertainty, we explore different channel switching patterns and different switching probabilities. In the case of multiple users, we analyze the probabilities of each user accessing channels with favorable channel conditions and the probability of collision. We also address a time-varying environment to identify the adaptive ability of the proposed framework. Additionally, we provide comparisons (in terms of both the average reward and time efficiency) between the proposed actor-critic deep reinforcement learning framework, Deep-Q network (DQN) based approach, random access, and the optimal policy when the channel dynamics are known.
\end{abstract}


\IEEEpeerreviewmaketitle

\begin{spacing}{1.5}

\section{Introduction}

The scarcity of spectral resources makes it challenging to satisfy the ever-growing demand for high-quality wireless communication services, and increases the importance to improve the spectrum utilization. Dynamic spectrum access, which enables users to proactively choose available channels, is one key approach to address this problem. However, dynamic spectrum access can be very challenging for instance in scenarios in which there is lack of prior information on the channel conditions or especially when the channel conditions in different frequency bands vary over time and multiple users dynamically access the channels simultaneously. Motivated by these considerations, we in this work propose a deep reinforcement learning based framework for dynamic multichannel access.

In particular, we consider an environment with $N$ correlated channels, and each channel is assumed to have two possible states: good or bad\footnote{Having more than two states can also be incorporated in the analysis and algorithms as done in Section \ref{sec:multi-user-priorities} with channels having ``excellent", ``good", and ``bad" states.}. The good state indicates better channel conditions and higher channel capacity, ensuring transmission success, while the bad state implies increased chances for transmission failure due to unfavorable channel conditions. It is assumed that the state of each channel can switch between good and bad, and this switching pattern can be modeled as a Markov chain with at most $2^N$ states. In order to successfully transmit their data, all users aim at selecting the good channels as frequently as possible. Since the channel switching pattern and other users' choices are unknown, each user can only try sensing or accessing different channels at each time and determine the pattern as much as possible based on its own observation. Here, we assume that users can receive a feedback in the channels they selected, and this channel feedback will indicate the channel conditions\footnote{We describe specific types of feedback that can indicate the channel condition in Section \ref{subsec:observation}.}. In this way, users learn if their selections lead to channels with good or bad states, and based on such previous experience, they predict the channel states in the next time period when they need to choose a channel, and increase the probability of choosing a channel in good state.

Since each user is only able to learn the states of channels selected by itself, the environment is partially observable to the users, making the channel selection problem a partially observable Markov decision process (POMDP). This is to say, to solve the dynamic spectrum access problem, an access policy that only depends on the user's individual information on the state of previously accessed channels after each time of sensing must be determined. However, in theory, POMDP problems are PSPACE-hard, and the increase in the number of states will lead to double-exponential growth in complexity. Hence, it is rather difficult to obtain the optimal solution. Conventionally, heuristic algorithms \cite{smith2004heuristic,szer2012maa} and Monte Carlo methods \cite{silver2010monte,thrun2000monte} have been used to find acceptable sub-optimal solutions in a reasonable duration of time. In both approaches, decisions are made based on previous exploration results.

In this paper, inspired by the effectiveness of reinforcement learning methods in exploring unknown environments \cite{Peters2005NaturalA,sutton1998reinforcement},  we investigate the use of deep reinforcement learning algorithms in solving the dynamic spectrum access problem. More specifically, we propose a deep actor-critic reinforcement learning framework for dynamic spectrum access, aiming at increasing the accuracy of channel selection with good states. Our main contributions in this work can be summarized as follows:
\begin{itemize}
	\item We propose an actor-critic deep reinforcement learning framework for dynamic multichannel access in a single-user scenario and show that this framework can work with a relatively larger number of channels than other deep reinforcement learning based approaches.
	\item We analyze the performance of the proposed framework and compare it with the deep Q-network (DQN) framework presented in \cite{wang2018deep}. Simulation results demonstrate that our proposed framework can achieve competitive performance in the case of $16$ channels, and better performance in the cases of $32$ and $64$ channels.
	\item We test the proposed approach in time-varying scenarios, and the results demonstrate the adaptive ability of actor-critic deep reinforcement learning. Also, our framework leads to significant benefits in terms of time/computational efficiency.
	\item We extent the actor-critic algorithm-based framework to a multi-agent framework to solve the dynamic multichannel selection problem in the multi-user model. And this purely distributed multi-agent framework can work without any additional information exchange between the users.
	\item We provide the channel selection accuracy for the case in which multiple users make their access decisions simultaneously, and compare with other algorithms (such as DQN, slotted ALOHA, and the optimal policy when the channel dynamics/patterns are known).
	
\end{itemize}

The remainder of the paper is organized as follows. We first review recent studies on the application of reinforcement learning algorithms in dynamic spectrum access problems in Section \ref{sec: related work}. Then, we introduce the system model for both single-user and multi-user scenarios in Section \ref{sec: system model} and formulate the corresponding optimization problems in Section \ref{sec: problem formulation}. In Section \ref{sec:algorithm}, we describe the proposed reinforcement learning algorithms and their workflows in detail. Subsequently, in Section \ref{sec: experiment}, we present the experimental/simulation settings and provide numerical results to evaluate the performance of the proposed framework. Finally, conclusions are drawn in Section \ref{sec : conclusion}.

\section{Related Work}\label{sec: related work}

The dynamic spectrum access problem has been extensively studied in the literature. For instance, the authors in \cite{hu2018full} provided a comprehensive survey on spectrum sharing technologies in cognitive radio networks with an outlook towards 5G. In the fixed spectrum assignment policy, a large portion of the assigned spectrum may be used sporadically where another portion of the spectrum can be congested. Allowing users to dynamically choose the available channels, the dynamic spectrum access technology is considered crucial to ensure that the limited spectral resources are allocated appropriately to satisfy the users' demand. For the correlated channel scenarios, the authors in \cite{zhao2007decentralized} developed an analytical framework for opportunistic spectrum access based on the theory of partially observable Markov decision processes (POMDPs). And for independent channels, the problem can be modeled as a restless multi-armed bandit (RMAB) process \cite{liu2008restless}.

Numerous studies have been conducted to find the spectrum access policies enabling users to effectively probe the channels. For instance, myopic policies were studied in \cite{zhao2008myopic} and \cite{ahmad2009optimality}, where the information on channels is collected through sufficient statistics and the user only senses the channel with the highest conditional probability. A stochastic game theory based policy was presented in \cite{xu2012opportunistic} and \cite{zheng2015stochastic}, where multiple users are in the system but each user can adjust its behavior based only on the individual information. In \cite{wang2017optimally}, a joint probing and accessing policy was proposed to allow the user to probe multiple channels at a time.

Inspired by the achievements of reinforcement learning in dynamic control problems, such as the game of Atari \cite{mnih2015human}, and AlphaGo \cite{silver2017mastering}, there has been increased interest in seeking reinforcement learning based solutions for problems in wireless communications. As summarized in \cite{luong2018applications} and \cite{zhang2018deep}, deep reinforcement learning algorithms have been applied in various wireless settings. For example, the authors in  \cite{xiao2018reinforcement} and \cite{ortiz2017reinforcement} investigate the use of Q-learning and SARSA (state-action-reward-state-action) reinforcement learning, respectively, in power control. The base stations' ON-OFF states are controlled by a deep Q-network (DQN) with the goal to improve the energy efficiency in \cite{li2018deep}. And authors of \cite{liu2018reinforcement} introduced the allocation of computational resources, and proposed a semi-MDP based optimal policy to schedule the cloud computing resources with the purpose to improve the system utility. Moreover, reinforcement learning is also used to perform joint optimizations. For example, the DQN was applied to seek optimal policies to jointly allocate the sub-bands and power in vehicle-to-vehicle communication \cite{ye2018deep}, and an actor-critic reinforcement learning framework was proposed to jointly solve the user scheduling, and subchannel and power allocation problem in order to maximize the energy efficiency \cite{wei2018user}.

As to the dynamic spectrum access, the control problem is generally modeled as either an MDP \cite{nassar2018reinforcement} \cite{yu2018deep} or a POMDP \cite{wang2018deep} depending on whether the environment is completely observable to the users or not. And, there are various different reinforcement learning algorithms being used in solving the spectrum access problems. Authors in \cite{dai2014online} proposed a continuous sampling and exploitation (CSE) online learning algorithm for an RMAB model. An application of Q-learning in the sensing order selection, in the presence of imperfect sensing, is presented in \cite{zhang2014model}.
Also, as a typical reinforcement learning framework, DQN has been applied in \cite{wang2018deep, naparstek2019deep, liu2018deep} for different purposes such as to improve the accuracy of selecting the channels in good condition, to  maximize the network utility, or to minimize the service blocking probability. Additionally, in order to solve the dynamic spectrum access problem in decentralized systems, different multi-agent reinforcement learning strategies are studied in \cite{naparstek2019deep,li2010multiagent,bkassiny2011distributed}. For instance, in \cite{naparstek2019deep}, the authors concentrated on a multi-user scenario in which transmission is successful only if a single user transmits over an accessed channel. The channels themselves do not inherently have time-varying states and correlations, and only collisions lead to transmission failures. And in \cite{yau2010enhancing}, a comparison between the single-agent reinforcement learning and multi-agent reinforcement learning is provided.

\section{System Model}\label{sec: system model}
In this work, we consider the dynamic multichannel access problem in which users dynamically select channels and learn the channel states. Below, we describe the system model in detail.

\subsection{Channel State Switching Patterns}

In the system we consider, there are $N$ correlated channels in total, and each channel has two possible states: the good channel state, which allows the user to transmit successfully, and the bad channel state, which will lead to transmission failure. We assume that the states of these channels are dynamically switching between good and bad. Since the channels are correlated, we can model the switching pattern of the channel states as a Markov chain, denoted as $\mathcal{P}$. In each time slot $t$, we denote the channel state as $\mathtt{X}_t = \{\x_{1,t}, \x_{2,t}, ..., \x_{N,t} \} $, where $N $ is the total number of channels, $\x_{i,t}$ stands for the state of the $i^{th}$ channel in time slot $t$.
And we assume that the channel state can only change at the beginning of each time slot and remains the same within the time slot. State transition probabilities at the beginning of each time slot can be described as follows: the probability that the channel state will change from current state to a different state in the Markov chain $\mathcal{P}$ is $p$; and the probability that the channel state will remain the same
is $(1-p)$.

\subsection{Users' Observations}\label{subsec:observation}

We assume that the channel switching pattern is unknown to the users. In order to successfully transmit their data, users have to deduce the channel switching pattern from their observations of the channels. Different mechanisms can be used to obtain such observations (or channel feedback). One approach is that the users send pilot or data signals over the selected channels and receive feedback from their corresponding receivers in the form of signal-to-interference-plus-noise ratios (SINRs). Or the users can tune to certain channels and determine the SINRs of signals received in those channels. In order to keep the analysis general in the paper, we assume that the users learn the conditions of the channels they have selected and accessed without explicitly detailing the particular mechanism. We assume that each user can only select $k$ channels to access, where $1 \le k < N$, and by accessing the selected channels, the user can learn the corresponding channel states while the states of all other channels that are not selected remain unknown to the user. Thus, from the user's perspective, choosing the channels in good states out of $N$ channels is a POMDP, in which the user aims to learn the pattern of variations in the channel states based on previous decisions. In the following two subsections, we describe the user observations initially in the case in which there is only one user in the system, and then in the case where there are multiple users trying to access the channels simultaneously.

\subsubsection{Single-User Scenario}

For the system with only one user, we denote the user's observation in time slot $t$ as $O_t = \{o_{1,t}, o_{2,t}, ..., o_{N,t} \}$, where $N$ is again the total number of channels, and $o_{i,t}$ stands for the user's observation of the $i^{\text{th}}$ channel, where $i = 1, 2, \dots, N $, in time slot $t$. We assume that when the user selects a channel to access, the state of the chosen channel is revealed to the user. In the single-user case, let us define the state of channel $i$, for $i = 1, 2, ..., N$, in time slot $t$ as
\begin{align}
\mathtt{x}_{i,t} = \begin{cases}
+1 \hspace{1cm} \text{if the $i^{\text{th}}$ channel is in good state in time slot $t$ }\\
-1 \hspace{1cm} \text{if the $i^{\text{th}}$ channel is in bad state in time slot $t$}
\end{cases}.
\end{align}
Now, for the user, the observation of each channel is
\begin{align}
o_{i,t}= \phi_{i,t} \, \x_{i,t} =
\begin{cases}
\x_{i,t}\hspace{1cm} \text{if the $i^{\text{th}}$ channel is selected in time slot $t$ }\\
0\hspace{1cm} \text{if the $i^{\text{th}}$ channel is not selected in time slot $t$}
\end{cases}
\end{align}
where $\phi_{i,t}$ is the indicator defined as
\begin{align}\label{eq:SU-SI}
\phi_{i,t}=
\begin{cases}
1  \hspace{1cm} \text{if the $i^{\text{th}}$ channel is selected in time slot $t$}\\
0  \hspace{1cm} \text{if the $i^{\text{th}}$ channel is not selected in time slot $t$}
\end{cases}.
\end{align}
As seen above, if a channel is not selected for access, its state is not known and we indicate the observation for those channels as zero.

\subsubsection{Multi-User Scenario}
For the system with $M > 1$ users, users make their own decisions to choose which channels to access and can only receive the feedback on the channel states corresponding to the selected channels. We assume that the users are not able to exchange information on their selections and observed channel states among themselves. Therefore, it is unavoidable that in some time slots, more than one user can choose/access the same channel. In these circumstances, even if the selected channels are in good state, the users may experience ``degraded" channels due to potential collisions. Taking this into account, we define the state of the $i^{\text{th} }$ channel in time slot $t$ as follows: 
\begin{align}
\x_{i,t}=
\begin{cases}
+1 \hspace{1.2cm} \text{if the $i^{\text{th}}$ channel is in good state and no collision occurs}\\
+1 \cdot d_{i,t} \hspace{0.3cm} \text{if the $i^{\text{th}}$ channel is in good state and collision occurs}\\
-1  \hspace{1.2cm} \text{if the $i^{\text{th}}$ channel is in bad state}
\end{cases}
\end{align}
where $d_{i,t} < 1$ is the discount factor for the good channels that are selected by more than one user. Note that this discount factor is introduced in order to discourage the users to access the same good channel in the same time slot so that collisions in channels with good states can be avoided as much as possible. In practice, different mechanisms can be employed for collision detection and different discount factor formulations can be used.

One approach is to define the discount factor $d_{i,t}$  to be proportional to $\frac{1}{m_i}$, where $m_i > 1$ is the number of users that have selected the $i^{\text{th}}$ channel. This choice can be justified as follows. As noted above, let us assume that the users receive SINR feedback from their corresponding receivers after accessing the selected channels. We denote the received power (after having experienced fading) when user $j$ accesses a good channel as $P^{good}_{r,j}$, while the received power when the user $j$ accesses a bad channel is indicated as $P^{bad}_{r,j}$. We further assume that $P^{good}_{r,j} \gg N_0 \gg P^{bad}_{r,j}$, where $N_0$ is the noise power. Now, we can choose two thresholds $\Gamma_1$ and $\Gamma_2$ with which the following inequalities with the received SINRs are satisfied:
\begin{equation}\label{Ieq:threshold}
\underbrace{\frac{P^{good}_{r,j}}{N_0}}_{\substack{\text{no interference/}\\ \text{good channel}}} > \Gamma_1 > \underbrace{\frac{P^{good}_{r,j}}{\sum\limits_{k \in \mathbb{I}, k \neq j} P^{good}_{r,k} + N_0}}_\text{interference/good channel} > \Gamma_2 >  \underbrace{\frac{P^{bad}_{r,j}}{N_0}}_{\substack{\text{no interference/}\\ \text{bad channel}}} > \underbrace{\frac{P^{bad}_{r,j}}{\sum\limits_{k \in \mathbb{I}, k \neq j} P^{bad}_{r,k} + N_0}}_\text{interference/bad channel}
\end{equation}
where the leftmost term in (\ref{Ieq:threshold}) is the signal-to-noise ratio (SNR) when user $j$ accesses a good channel and there are no other users in this channel (and hence no interference). Note that since we assume that the received power in a good channel satisfies $P^{good}_{r,j} \gg N_0$, we have $\frac{P^{good}_{r,j}}{N_0} \gg 1$. On the other hand, if multiple users access a good channel and a collision occurs, the SINR of user $j$ becomes $\frac{P^{good}_{r,j}}{\sum\limits_{k \in \mathbb{I}, k \neq j} P^{good}_{r,k} + N_0}$, where $\mathbb{I}$ denotes the index set of interfering users. Assuming that the received powers of different users at the same receiver are comparable, we have
\begin{gather} \label{eq:discount-factor-justification}
\frac{P^{good}_{r,j}}{\sum\limits_{k \in \mathbb{I}, k \neq j} P^{good}_{r,k} + N_0} \approx \frac{P^{good}_{r,j}}{\sum\limits_{k \in \mathbb{I}, k \neq j} P^{good}_{r,k}} \approx \frac{1}{m-1}
\end{gather}
where $m$ is the number of users that select the good channel in the given time slot, and the first approximation is due to received powers being much larger than noise power. The second approximation in (\ref{eq:discount-factor-justification}) (which is due to received power levels being comparable) provides a justification for choosing the discount factor $d_{i,t}$ to be proportional to $\frac{1}{m}$. As to the cases in which users select bad channels, since the received power levels are small (e.g., because of potentially strong attenuation in the channel) and the noise is dominant, SNR and SINRs levels will be small. Finally, we note that if there exists thresholds $\Gamma_1$ and $\Gamma_2$ satisfying the inequalities in (\ref{Ieq:threshold}), comparisons with these thresholds would serve as one approach to identify channel states and recognize collisions via SNR/SINR feedback.

Now, we define the observation of user $j$ in time slot $t$ as  $O_{j,t} = \{o_{j,1,t}, ..., o_{j,i,t}, ..., o_{j,N,t} \}$, where $j$ is the user index and $i$ is the index of the channel. Similar to the single-user scenario, the user $j$'s observation of the $i^{\text{th} }$ channel is
\begin{align}
o_{j,i,t}= \phi_{i,j,t} \, \x_{i,t} =
\begin{cases}
\x_{i,t}\hspace{1cm} \text{if the $i^{\text{th}}$ channel is selected in time slot $t$ }\\
0\hspace{1cm} \text{if the $i^{\text{th}}$ channel is not selected in time slot $t$}
\end{cases}
\end{align}
where $\phi_{i,j,t}$ is an indicator defined as
\begin{align} \label{eq:MU-SI}
\phi_{i,j,t}=
\begin{cases}
1  \hspace{0.5cm} \text{if the $i^{\text{th}}$ channel is selected by user $j$ in time slot $t$}\\
0  \hspace{0.5cm} \text{if the $i^{\text{th}}$ channel is not selected by user $j$ in time slot $t$}
\end{cases}.
\end{align}
Similarly as in the single-user case, the channel states are revealed only for the selected/accessed channels. For the channels that are not selected for access, the observation is set to zero.

\subsubsection{Users' Action Space}
As we noted before, the users can only select $k$ channels to access in each time slot, where $1 \le k < N$. We consider a discrete action space $\mathcal{A} = \{a_1, a_2, \dots, a_{\mathcal{D} } \}$, where $\mathcal{D}$ is the total number of valid actions. Each valid action in the action space describes the $k$ indices of the channels that will be accessed. So, for a specific value of $k$, we have the number of actions equal to $N\choose{k}$. For example, if $k=1$, each action $a_i$, $i = 1,2,...,\mathcal{D} = N$, corresponds to accessing channel $i$; while if $k = 2$, each valid action $a_i$, $i = 1,2,...,\mathcal{D}$, can be described by the indices of the two chosen channels. Hence, in each time slot, the user will pick one action from the action space $\mathcal{A}$, access the corresponding $k$ channels, and the condition of the chosen channels will be revealed.

\section{Multichannel Access Problem Formulation}\label{sec: problem formulation}

In this section, we formulate the dynamic multichannel access problem based on the channel access mechanisms and the corresponding rewards. To learn the channel switching pattern, we propose an actor-critic algorithm based deep reinforcement learning framework, which works as an agent to make channel selection decisions for the user. In this framework, the agent obtains the user's observation of the channels and makes channel access decisions based on the observation, and subsequently receives the feedback from the channels, and updates the decision policy.

\subsection{Single-User Scenario}
We first consider the case in which there is only one user in the system. The reward $r_{i,t}$ obtained when the $i^{\text{th}}$ channel is selected/accessed by the user in time slot $t$ is defined as follows:
\begin{align}\label{eq:reward-singleuser}
r_{i,t} = \x_{i,t} = \begin{cases}
+1 \hspace{1cm} \text{if the $i^{\text{th}}$ channel is in good state in time slot $t$ }\\
-1 \hspace{1cm} \text{if the $i^{\text{th}}$ channel is in bad state in time slot $t$}
\end{cases}.
\end{align}
Since the user aims to select good channels as much as possible to ensure frequent successful transmissions, the agent is designed to find a policy $\pi$ (which is a mapping from the observation space $\mathcal{O}$ to the action space $\mathcal{A}$) that maximizes the long-term expected reward $R$ of channel access decisions:
\begin{equation}
\pi^* = \arg \max_{\pi} R
\end{equation}
where $\pi^*$ denotes the optimal decision policy, and in a finite time duration $T$, we express $R$ as
\begin{equation} \label{eq:avgreward-singleuser}
R = \frac{1}{T} \sum_{t = 1}^{T} \sum_{i = 1}^{N} \phi_{i,t} \, r_{i,t}
\end{equation}
where $\phi_{i,t}$ is the indicator function defined in \eqref{eq:SU-SI}.
Now, the problem can be formulated as
\begin{align}
\textbf{P1:}\hspace{1cm}&\underset{\{\phi_{i,t}\}}{\text{Maximize}}\hspace{1.2cm} R\\
&\text{Subject to} \hspace{0.7cm} \sum_{i=1}^{N} \phi_{i,t} = k
\end{align}
where $k$ is the number of channels that the user can select in each time slot, and according to the definition of $R$, we have $R \in [-k, k]$.

\subsection{Multi-User Scenario}
In this case, we assume that there are multiple users, and each user can independently select a channel to access without knowing other users' actions. Thus, each user will employ a separate actor-critic reinforcement learning agent. Since each user has the goal to choose good channels as frequently as possible, the agent of user $j$ is required to find a policy $\pi_j$ for $j = 1,2, \dots, M$ (mapping the observation space $\mathcal{O}_j$ to the action space $\mathcal{A}$) that maximizes the long-term expected reward $R_{j}$ of the channel access decisions for user $j$:
\begin{equation}
\pi^*_j = \arg \max_{\pi} R_j.
\end{equation}
Similarly as in the single-user case, $\pi^*_j$ denotes the optimal decision policy for user $j$, and in a finite time duration $T$, we express $R_j$ as
\begin{equation}\label{eq:avgreward-multiuser}
R_j = \frac{1}{T} \sum_{t = 1}^{T} \sum_{i = 1}^{N} \phi_{i,j,t} \, r_{i,t}
\end{equation}
where $\phi_{i,j,t}$ is an indicator defined in (\ref{eq:MU-SI}) and the reward $r_{i,t}$ obtained when user $j$ accesses the $i^{\text{th}}$ channel in time slot $t$ is
\begin{align}\label{eq:reward-multiuser}
r_{i,t} = \x_{i,t} =
\begin{cases}
+1 \hspace{1.2cm} \text{if the $i^{\text{th}}$ channel is in good state and no collision occurs}\\
+1 \cdot d_{i,t} \hspace{0.3cm} \text{if the $i^{\text{th}}$ channel is in good state and collision occurs}\\
-1  \hspace{1.2cm} \text{if the $i^{\text{th}}$ channel is in bad state}
\end{cases}.
\end{align}

Hence, the optimization problem for user $j$ for $j = 1, 2,\dots, M$ can be formulated as
\begin{align}
\textbf{P2:}\hspace{1cm}&\underset{\{\phi_{i,j,t}\}}{\text{Maximize}}\hspace{1.2cm} R_j\\
&\text{Subject to} \hspace{0.7cm} \sum_{i=1}^{N} \phi_{i,j,t} = k.
\end{align}

The formulation of each user's optimization problem is similar to that in the single-user case. However, the optimal solution in the multi-user scenario should find the channels in good condition and also avoid collisions at the same time. This means that the agent needs to learn both the channel switching pattern and the other users' channel selection pattern from the channel feedback. When there are not enough channels in good state, users compete for such limited number of good channels. On the other hand, if a sufficient number of good channels exists simultaneously, each user can potentially access a good channel without experiencing a collision.


\section{Actor-Critic Reinforcement Learning Framework} \label{sec:algorithm}

In this section, we describe the proposed actor-critic deep reinforcement learning framework for dynamic multichannel access and develop algorithms for both single- and multi-user cases.

\subsection{Actor-Critic Agent's Observation Space, Actions, and Rewards}
We first introduce the relevant definitions within the actor-critic framework.

\begin{figure}
	\centering
	 \includegraphics[width=.7\linewidth]{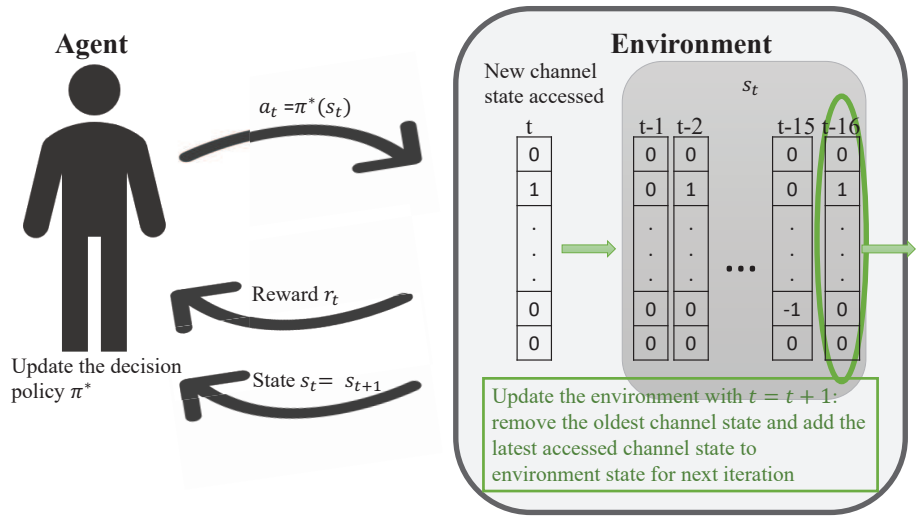}
	\caption{In reinforcement learning, the agent constantly observes the environment and makes a decision. The decision will be executed and the corresponding feedback will be used to update the policy.}\label{fig:system-figures2}
\end{figure}

\emph{Channel State and Agent's Observation:} The channel state is varying as described by a Markov chain and it is  a part of the environment, which is unknown to the agent. Therefore, the agent can only take its own observation space $\mathcal{O}$ as the input to the actor-critic framework. The agent can only access the chosen channels in each iteration, and observe the reward that depends on the state of the chosen channels. As defined in the previous section, in time slot $t$ the user observation $O_t$ (or $O_{j,t}$ for multi-user scenario) is a sparse matrix with only $k$ nonzero elements in each column (representing the observation vector at any given time), where $k$ is the number of channels that are selected to be accessed in each time slot. The users will learn on the basis of their previous experiences. We assume the agent keeps an observation space $\mathcal{O}$ that consists of the most recent $\Omega$ observations $O_t$. The observation space is initialized as an all-zero $N \times \Omega$ matrix, and at each time $t$, the latest observation $O_t$ will be added to the observation space, and oldest observation $O_{t-\Omega}$ will be removed. The updated observation space $\mathcal{O}$ at time $t+1$ is denoted as $\mathcal{O}_{t+1} = \{O_{t}; O_{t-1}; ...; O_{t-(\Omega-1)} \}$. For instance, Fig. \ref{fig:system-figures2} depicts a scenario in which $\Omega = 16$ and $k = 1$.

\emph{Action:} The agent scores all possible actions in the action space $\mathcal{A}$ based on the user's observation, and the action with the highest score will be chosen. In our setting, the action indicates which channel or channels to access.

\emph{Reward:} The reward is received when the action is executed, meaning that the agent chooses a channel and gets direct feedback from the environment. The reward is defined based on the condition of the chosen channel, as formulated in the optimization problems \textbf{P1} and \textbf{P2}.

In addition to the average reward formulations given in (\ref{eq:avgreward-singleuser}) and (\ref{eq:avgreward-multiuser}), we also define here the reward received by the agent in time slot $t$ since this will be used in the actor-critic reinforcement learning algorithm. In particular, in the single-user scenario, the reward in time slot $t$ is
\begin{equation}\label{eq:reward-at-time-t-singleuser}
R_t = \sum_{i = 1}^{N} \phi_{i,t} \, r_{i,t}
\end{equation}
where $ \phi_{i,t}$ is given in (\ref{eq:SU-SI}) and $r_{i,t}$ is given in (\ref{eq:reward-singleuser}).

In the multi-user case, the reward for agent $j$ in time slot $t$ can be expressed as
\begin{equation}\label{eq:reward-at-time-t-multiuser}
R_{j,t} = \sum_{i = 1}^{N} \phi_{i,j,t} \, r_{i,t}
\end{equation}
where $\phi_{i,j,t}$ is the channel selection indicator for agent $j$ as defined in (\ref{eq:MU-SI}), and $r_{i,t}$ is provided in (\ref{eq:reward-multiuser}).

\subsection{Algorithm Overview}\label{sub:algorithm}
\begin{figure}[tb]
	\centering
		\includegraphics[width=0.7\linewidth]{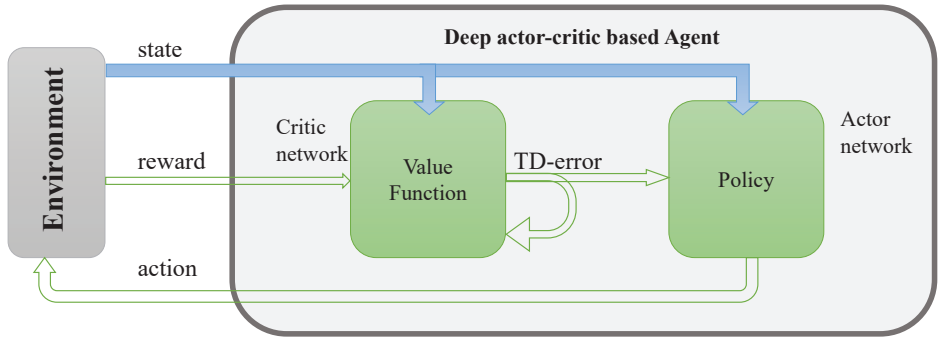}
	\caption{Structure of the actor-critic deep reinforcement learning agent}
	\label{Fig.structure}
\end{figure}

In this subsection, we describe the architecture of the actor-critic algorithm. The actor-critic architecture consists of two neural networks: actor and critic. In our model, the actor neural network is parameterized by $\theta$, and the critic neural network is parameterized by $\mu$. The structure of the actor-critic deep reinforcement learning agent is depicted in Fig. \ref{Fig.structure}

\emph{Actor:} The actor is employed to explore a policy $\pi$, that maps the agent's observation $\mathcal{O}$ to the action space $\mathcal{A}$:
\begin{equation}
\pi_{\theta}(\mathcal{O}) : \mathcal{O} \rightarrow \mathcal{A}
\end{equation}
So the mapping policy $\pi_{\theta}(\mathcal{O})$ is a function of the observation $\mathcal{O}$ and is parameterized by $\theta$. And the chosen actor can be denoted as
\begin{equation}
a = \pi_{\theta}(\mathcal{O})
\end{equation}
where we have $a \in \mathcal{A}$. Since the action space is discrete, we use softmax function at the output layer of the actor network so that we can obtain the scores of each actions. The scores sum up to $1$ and can be regarded as the probabilities to obtain a good reward by choosing the corresponding actions.

\emph{Critic:} The critic is employed to estimate the value function $V(\mathcal{O})$. At time instant $t$, when the action $a_t$ is chosen by the actor network, the agent will execute it in the environment and send the current observation $\mathcal{O}_t$ along with the feedback from the environment to the critic. The feedback includes the reward $r_t$ and the next time instant observation $\mathcal{O}_{t+1}$. Then, the critic calculates the TD (Temporal Difference) error:
\begin{equation} \label{eq:TDerror}
\delta_{t} = R_t + \gamma V_{\mu}(\mathcal{O}_{t+1}) - V_{\mu}(\mathcal{O}_t)
\end{equation}
where $\gamma \in (0,1)$ is the discount factor \footnote{We note that the reward $R_t$ in (\ref{eq:TDerror}) is given by (\ref{eq:reward-at-time-t-singleuser}) in the single-user case, and is equal to $R_{j,t}$ in (\ref{eq:reward-at-time-t-multiuser}) when user/agent $j$ is considered in the multi-user scenario.}.

\emph{Update:} The critic is updated by minimizing the least squares temporal difference (LSTD):
\begin{equation}
V^* = \arg \min_{V_{\mu}} (\delta_{t} )^2
\end{equation}
where $V^*$ denotes the optimal value function.

The actor is updated by policy gradient. Here, we use the TD error to compute the policy gradient\footnote{In (\ref{eq:policygradient}), policy gradient is denoted by $\nabla_{\theta} J(\theta)$ where $J(\theta)$ stands for the policy objective function, which is generally formulated as the statistical average of the reward.}:
\begin{equation}\label{eq:policygradient}
\nabla_{\theta} J(\theta) = E_{\pi_{\theta} } [ \nabla_{\theta} \log \pi_\theta(\mathcal{O}, a)  \delta_{t} ]
\end{equation}
where $\pi_\theta(\mathcal{O}, a)$ denotes the score of action $a$ under the current policy. Then, the weighted difference of parameters in the actor at time $t$ can be denoted as $\Delta\theta_{t} = \alpha \nabla_{\theta_t} \log \pi_{\theta_t}(\mathcal{O}_t, a_t) \delta_{t}$, where $\alpha \in (0,1)$ is the learning rate. And the actor network $i$ can be updated using the gradient decent method:
\begin{equation}
	\theta_{t+1} = \theta_t + \alpha \nabla_{\theta_t} \log \pi_{\theta_t}(\mathcal{O}_t, a_t) \delta_{t}.
\end{equation}

\subsection{Workflow for a Single User}
In the case of a single user, there is only one actor-critic network employed as the agent to dynamically select channels. At the beginning of time slot $t$, the agent will collect the latest $\Omega$ observations of channels and the observation space is denoted as $\mathcal{O}_t$. Then the actor network will choose $k$ channels according to the decision policy, i.e., the action with highest score will be selected. Next, the channel reward will be sent from every chosen channel. Based on the reward, the current observation space $\mathcal{O}_t$ and the observation space for the next time slot $\mathcal{O}_{t+1}$, the critic network calculates the TD-error. And finally the critic and actor networks will be updated based on the TD-error.

The full framework is provided in Algorithm \ref{alg:AC-SU} below.

\begin{algorithm}
	\caption{Actor-Critic Deep Reinforcement Learning Algorithm for Single-User Dynamic Multichannel Access}
	\label{alg:AC-SU}
	\begin{algorithmic}
		\State Initialize the critic network $V_{\mu}(\mathcal{O} )$ and the actor $\pi_{\theta}(\mathcal{O})$, parameterized by $\mu$ and $\theta$ respectively.
		\State The environment initializes the state of each channel $\mathtt{X}$.
		\State The agent initializes its observation as all zero matrix $\mathcal{O}_0$		
		\For{$t = 0,T$}
		
		\State With the observation, the agent selects $k$ channels according to the decision policy $a_t = \pi(\mathcal{O}_t| \theta)$ w.r.t. the current policy
		\State Agent accesses the chosen channels and receives the reward $R_t$ based on the channel state.
		\State Based on the reward, the new observation of channels $O_t$ will be added to the observation space for the next time slot $\mathcal{O}_{t+1}$
		
		\State Critic calculates the TD error: $ \delta_{t} = R_t + \gamma V(\mathcal{O}_{t+1}) - V(\mathcal{O}_t) $
		\State Update the critic by minimizing the loss: $\mathcal{L}(\mathcal{O}_t, a_t) = (\delta_{t} )^2$
		\State Update the actor policy by maximizing the action value: $\Delta\theta_t = \alpha \nabla_{\theta_t} \log \pi_{\theta_t}(\mathcal{O}_t, a_t) \delta_{t}$, $\alpha \in (0,1)$.	
		\State Update the observation $\mathcal{O}_t = \mathcal{O}_{t+1}$.	
		\State Update the channel state $\mathtt{X}$.
		
		\EndFor
	\end{algorithmic}
\end{algorithm}

\subsection{Workflow for Multiple Users}
In the case that multiple users access the channels simultaneously, we assume that all access decisions and actions are completed at the same time. At the beginning of the time slot $t$, agent $j$ for  $j = 1, 2, \dots, M$ collects the corresponding user's observation $\mathcal{O}_{j,t}$ of all channels. Then, each user will select the action with highest score according to its own decision policy. Next, the agents receive the rewards from their chosen channels simultaneously. Based on their own rewards and observations, critic networks will calculate the corresponding TD-error to update the critic and actor networks, respectively.

The full framework is provided in Algorithm \ref{alg:AC-MU} below.
\begin{algorithm}
	\caption{Actor-Critic Deep Reinforcement Learning Algorithm for Multi-User Dynamic Multichannel Access}
	\label{alg:AC-MU}
	\begin{algorithmic}
		\State Initialize the critic network $V_{\mu_j}(\mathcal{O}_j )$ and the actor $\pi_{\theta_j}(\mathcal{O}_j)$ for user $j$, parameterized by $\mu_j$ and $\theta_j$ respectively, with $j = 1,2, \dots, M$.
		\State The environment initializes the state of each channel $\mathtt{X}$.
		\State The agent $j$ initializes its observation as all zero matrix $\mathcal{O}_{j,0}$, $j =  1,2, \dots, M$.	
		\For{$t = 0,T$}
		\For{$j = 1,M$}
		
		\State With the observation, the agent selects an action $a_{j,t} = \pi(\mathcal{O}_{j,t}| \theta_j)$ w.r.t. the current policy
		\EndFor
		\State Agents start accessing the chosen channels simultaneously, and every agent receives the corresponding reward $R_{j,t}$ based on the channel state and the access collisions.
		\For{$j = 1,M$}
		\State Based on the reward, agent $j$ adds the new observation of channels $O_{j,t}$ to the observation space for the next time slot $\mathcal{O}_{j,t+1}$
		\State Every critic calculates the corresponding TD error: $ \delta_{j,t} = R_{j,t} + \gamma V(\mathcal{O}_{j,t+1}) - V(\mathcal{O}_{j,t}) $
		\State Update the critic by minimizing the loss: $\mathcal{L}(\mathcal{O}_{j,t}, a_{j,t}) = (\delta_{j,t} )^2$
		\State Update the actor policy by maximizing the action value: $\Delta\theta_{j,t} = \alpha \nabla_{\theta_{j,t}} \log \pi_{\theta_{j,t}}(\mathcal{O}_{j,t}, a_{j,t})  \delta^{\pi_{\theta_{j,t}}}$, $\alpha \in (0,1)$.
		\State Update the observation $\mathcal{O}_{j,t} = \mathcal{O}_{j,t+1}$ for every user.
		\EndFor
			
		\State Update the channel state $\mathtt{X}$.
		
		\EndFor
	\end{algorithmic}
\end{algorithm}

\section{Experiments and Numerical Results}\label{sec: experiment}

In this section, we initially describe the simulation setting and then evaluate the performance of the proposed actor-critic framework via numerical results, and provide comparisons with random channel access, the DQN based framework proposed in \cite{wang2018deep} and also the optimal policy under the assumption that the channel switching patterns are known.
\vspace{-.1cm}

\subsection{Simulation Setting}
In our implementation, the design of the agent for a single user and that of the agent for each user in the multi-user case are similar. The agent consists of two neural networks: actor and critic. Each of the two networks has two layers. For the actor, which scores all actions in the action space, the first layer has $200$ neurons with ReLU as the activation function, and second layer has $N$ neurons with Softmax as the activation function, where $N$ is the total number of channels. For the critic, which computes the value of the chosen action, the first layer has $200$ neurons with ReLU as the activation function, and second layer has $1$ neuron. Especially, since the critic will evaluate the decision made by the actor, the learning rate of the actor network should be smaller than that of the critic network to make the actor network converge slower than critic network. Here, we set the learning rate of the critic network as $0.0005$, and the learning rate of the actor network as $0.0001$. To ensure stability, both learning rates will decay exponentially with the decay rate $0.95$ for every $250000$ time slots.


\subsection{Average Reward in the Single-User Case}
In this section, we present the results for the single-user model, and compare our framework with the DQN framework, random access, Whittle index heuristic, and also the optimal decision policy under the assumption that the channel switching pattern is known to the user.

\emph{DQN framework} proposed in \cite{wang2018deep} consists of two hidden layers and maintains a replay memory with a size of $1,000,000$. To update the network, the DQN framework will replay a minibatch of $32$ samples extracted from the memory.

In the \emph{random access policy}, there is no learning and users randomly select channels at the beginning of each time slot, and all channels will be accessed with the same probability.

In \emph{Whittle index heuristic} \cite{liu2010indexability}, all the channels are treated as independent and the transition probability matrix of each channel is assumed to be obtained by observing the channels separately over a certain period (e.g. 10,000 iterations) in advance. Then, the transition probability matrices are used to update the belief vector for final decision making. More details can be found in \cite{liu2010indexability}.

We also consider the \emph{optimal policy} \cite[Theorem 1]{wang2018deep} assuming that the channel dynamics is known to the user. The user accesses a channel at the beginning. Then, according to the policy, for instance when the channel state switching probability is greater that $0.5$, if the user selects a good channel at time $t$, the user will choose a channel in the next activated subset of channels according to the known pattern in the next time slot. On the other hand, if the user selects a bad channel at time $t$, the user will stay at the chosen channel in the next time slot. The reverse strategy is employed if the switching probability is less than 0.5.

\subsubsection{Single Good Channel}
In this experiment, we consider different number of channels, i.e., $N = \{16, 32, 64\}$, and only one channel is in good state in each time slot. To evaluate the performance, we calculate the expected reward $R$ with different  Markov chains $\mathcal{P}$. To define a Markov chain for the channel distribution, we need to specify the channel states in order and the state switching probabilities. We assume that, for each state, the probability that current state will transfer to another state is $p$, and the probability that the current state will be kept is $1-p$. Our experiments were conducted in two cases:

\begin{figure}
\centering
\includegraphics[width=0.7\linewidth]{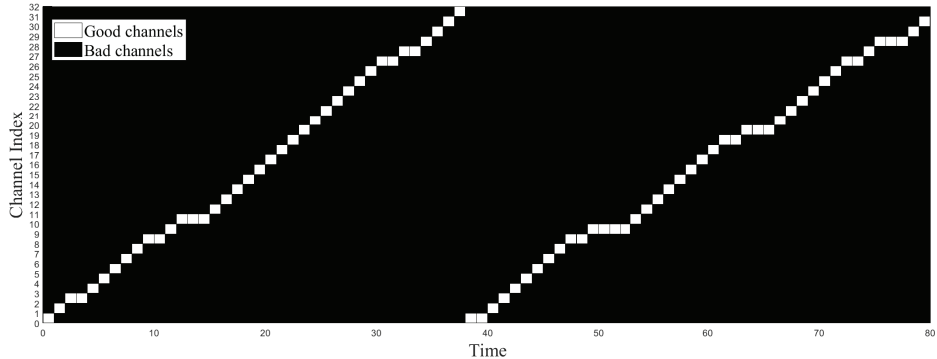}
\caption{Round-robin switching pattern when only one of the 32 channels is in good condition and the switching probability is $p = 0.75$. The channel in good state at a given time is indicated by a white square.}
\label{fig:RoundRobinPattern_32}
\end{figure}

\textbf{Round-Robin Switching Scenario:}
In this experiment, we assume that the index of the only  good channel switches from $1$ to $N$ according to a round-robin scheduling and we assume the user can only access to one channel at a time. Then we vary the switching probabilities as $p = \{0.75, 0.80, 0.85, 0.90, 0.95\}$. This round-robin pattern with switching probability $p = 0.75$ is depicted in Fig. \ref{fig:RoundRobinPattern_32}, where the channel with the good state is indicated with a white square at the corresponding channel index value at a given time. Since the probability $p = 0.75$ is relatively high, we have an increasing staircase pattern (indicating channels with good state changing from one to the next) more frequently than the flat pattern that occurs when the same channel stays in good state in multiple time slots.

\begin{figure}
	
	\begin{minipage}{.5\linewidth}
		\centering
		\subfloat[16 channels]{\label{fig:singleaction_A}\includegraphics[width=1.15\textwidth]{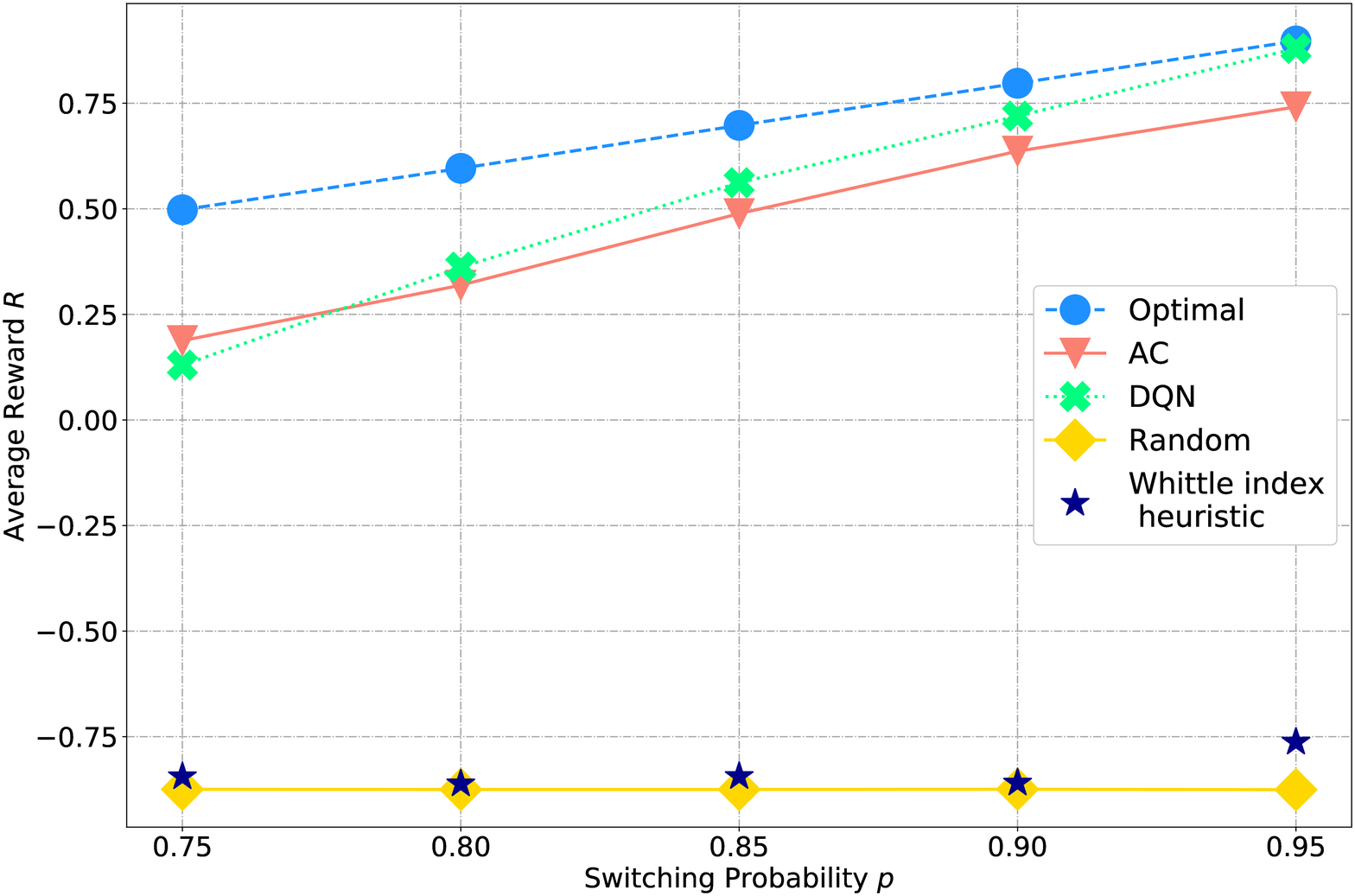}}
	\end{minipage}%
	\begin{minipage}{.5\linewidth}
		\centering
		\subfloat[32 channels]{\label{fig:singleaction_B}\includegraphics[width=1.15\textwidth]{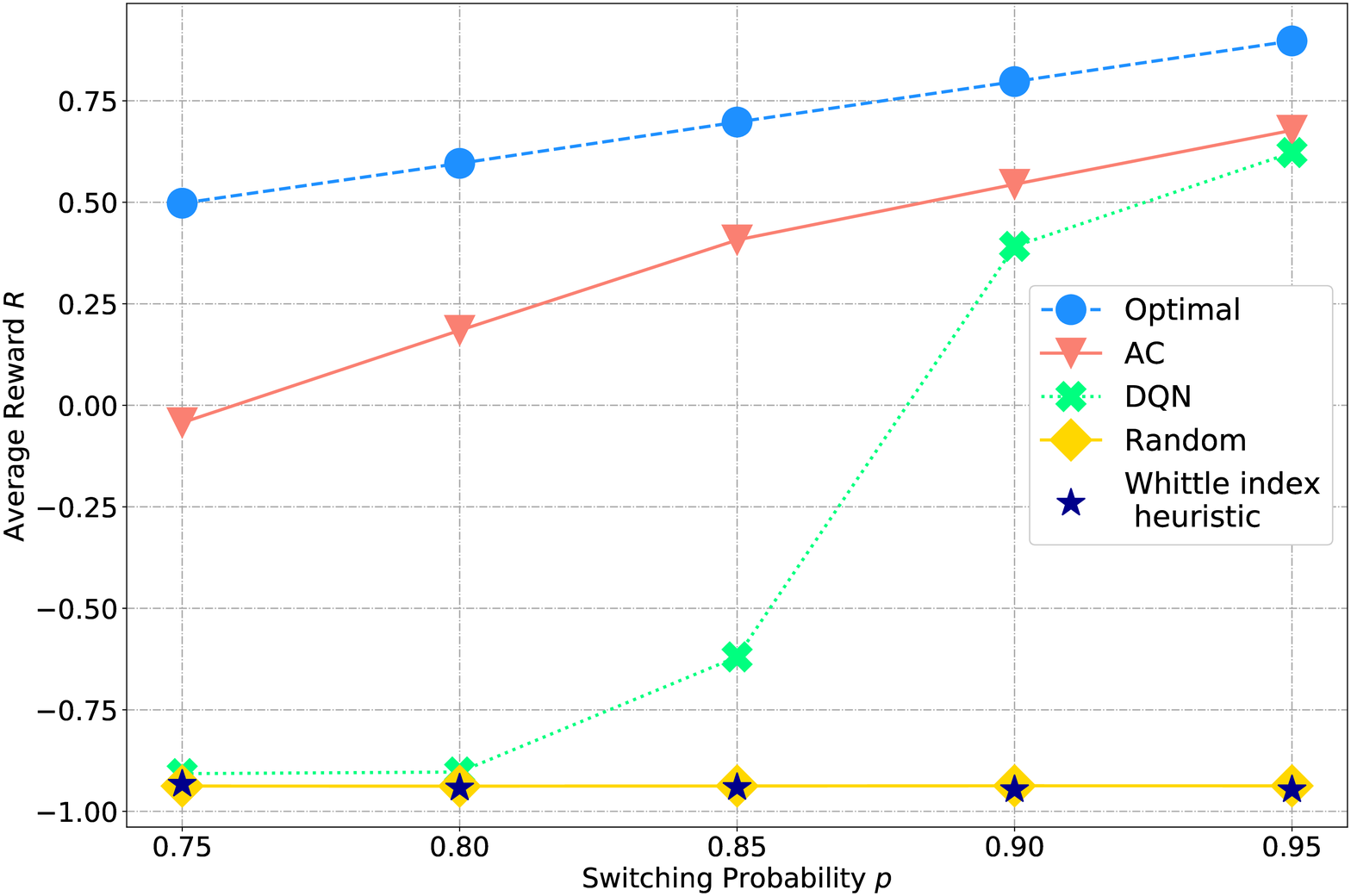}}
	\end{minipage}\par\medskip
	\centering
	\subfloat[64 channels]{\label{fig:singleaction_C}\includegraphics[width=0.57\textwidth]{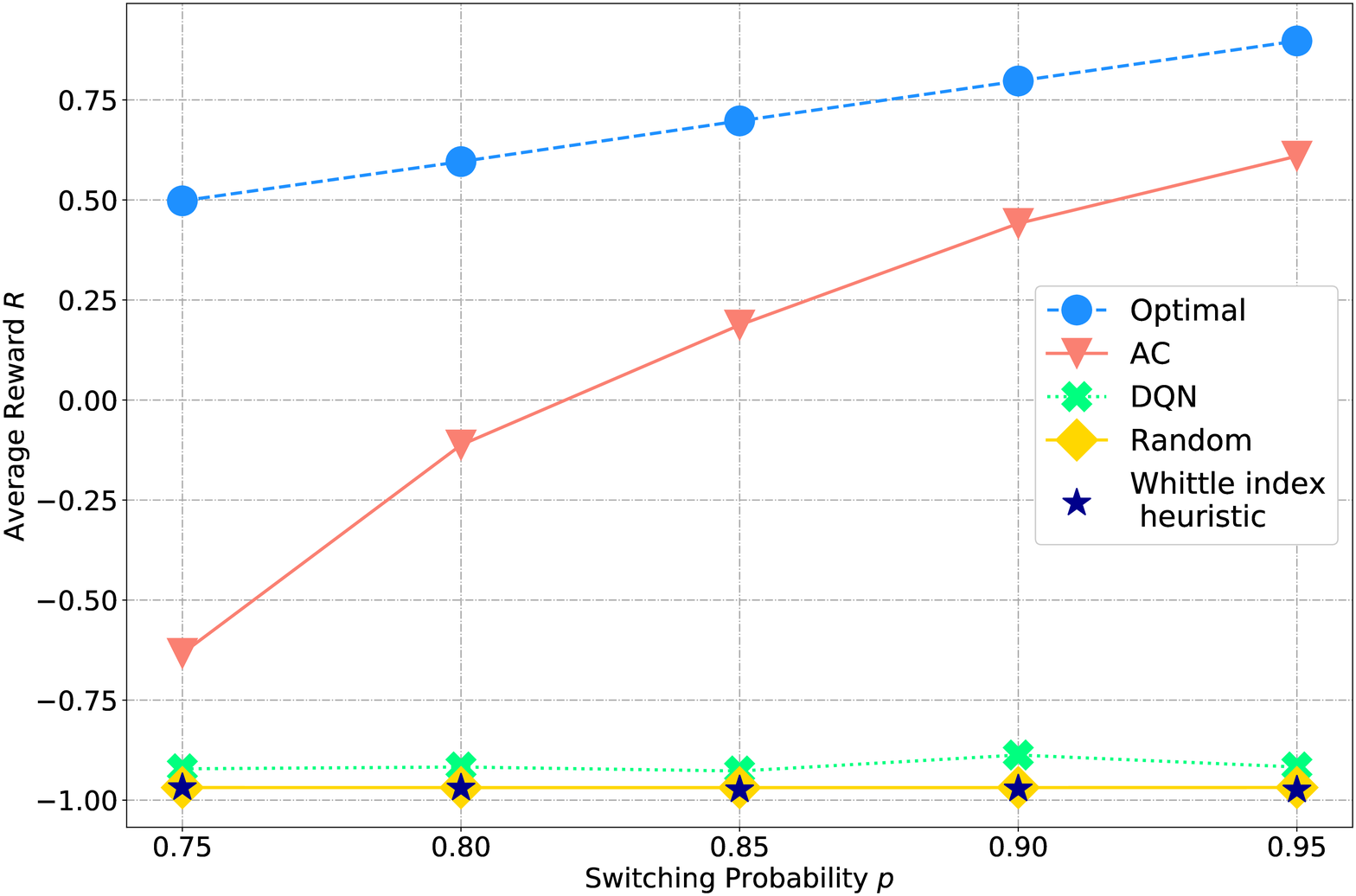}}
	
	\caption{Average reward vs. switching probability. We consider $16$, $32$, $64$ channels cases with the switching probability varies as $p = \{0.75, 0.80, 0.85, 0.90, 0.95\}$}
	\label{fig:singleaction}
\end{figure}

We compare our actor-critic (AC) policy with DQN, random access, Whittle index heuristic and optimal policy in terms of the average reward.  Figs. \ref{fig:singleaction}(a), \ref{fig:singleaction}(b), and \ref{fig:singleaction}(c) provide the average rewards of different policies for $N = 16, 32$ and $64$ channels, respectively. In all subfigures, we notice that the average rewards of the optimal policy are identical because the channel pattern is assumed to be known in this case, and hence the increase in the number of channels makes no influence on the policy performance. Optimal policy expectedly leads to the highest average rewards. On the other hand, performance curves achieved by Whittle index heuristic and the random access policy are very low, demonstrating the inadequacy of these strategies.

More interesting and competitive performances are displayed by the proposed actor-critic policy and DQN policy. We observe in Fig. \ref{fig:singleaction}(a) that when $N = 16$, DQN achieves a slightly higher reward than the actor-critic policy as the switching probability increases. However, the actor-critic policy performs consistently for different number of channels and outperforms DQN when the nunber of channels is increased to $N = 32$ and $N = 64$.
In particular, the DQN framework cannot handle the case of $64$ channels for any switching probability; and for the case of 32 channels, the DQN framework achieves negative rewards when $p \le 0.85$.
Hence, the proposed actor-critic framework is more suitable when the number of channels is relatively large.

As to the overall tendency in the actor-critic and DQN performances, except for the DQN curve in the case of $N = 64$ channels, we can observe increasing average reward as the switching probability increases and the gap between different cases diminishes. Note that $p$ denotes the probability of switching between states, and hence higher switching probability will decrease the uncertainty and will make it easier for the agent to learn the policy. Comparing the performances at the relatively low value of $p = 0.75$, the actor-critic framework demonstrates benefits for all $N$ values, and therefore it has higher tolerance against uncertainty.


\textbf{Arbitrary Switching Scenario:}

\begin{figure}
	\centering
	\includegraphics[width=0.9\linewidth]{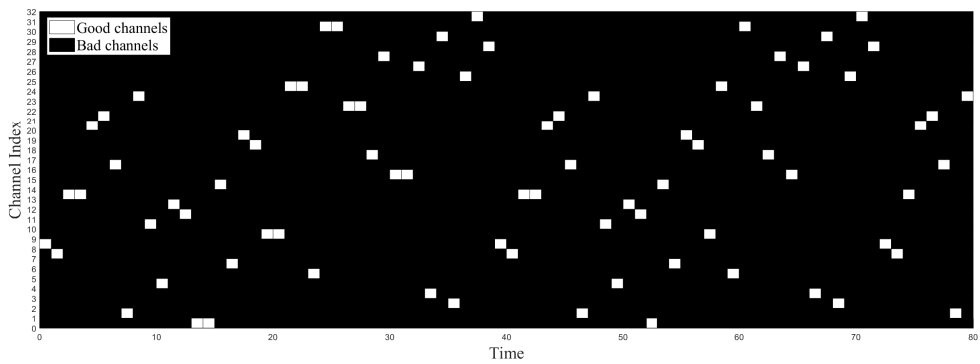}
	\caption{A switching pattern when only one of the 32 channels is in good condition at a given time, with a switching probability $p = 0.9$}
	\label{fig:AbitraryPattern_32}
\end{figure}

\begin{figure}
	\centering
	\includegraphics[width=.8\linewidth]{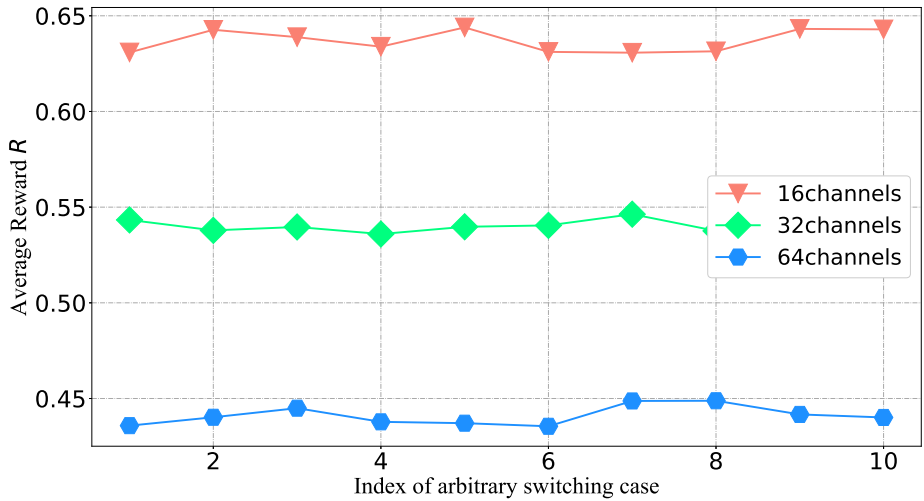}
	\caption{The average reward for different arbitrary switching orders}
	\label{fig:arbitrary}
\end{figure}

In the round-robin switching scenario, the channel states switch according to a specific scheduling model. However, this information is unknown to the actor-critic agent, and of course is not being used in the process to find a channel access policy. Moreover, the actor-critic algorithm was proposed as a model-free algorithm. To demonstrate the performance of the proposed framework in a model-free environment, we in this experiment, fix the switching probability $p$ at $0.9$ and test the framework with $10$ different arbitrary switching orders (i.e., $10$ different permutations of $N$ channels). One such switching pattern in the case of $N = 32$ channels is depicted in Fig. \ref{fig:AbitraryPattern_32}.

Fig. \ref{fig:arbitrary} plots the performance in the cases of $16$, $32$ and $64$ channels with $10$ randomly generated arbitrary switching orders. Still, the user is allowed to access one channel at a time. For any given number of channels, the average reward varies only slightly across different switching cases, showing that our proposed framework can work in a model-free environment. Since we have shown that the switching order will not affect the agent's performance, we assume a round robin switching scheduling in all the following experiments.

\begin{figure}
	\centering
	\includegraphics[width=0.9\linewidth]{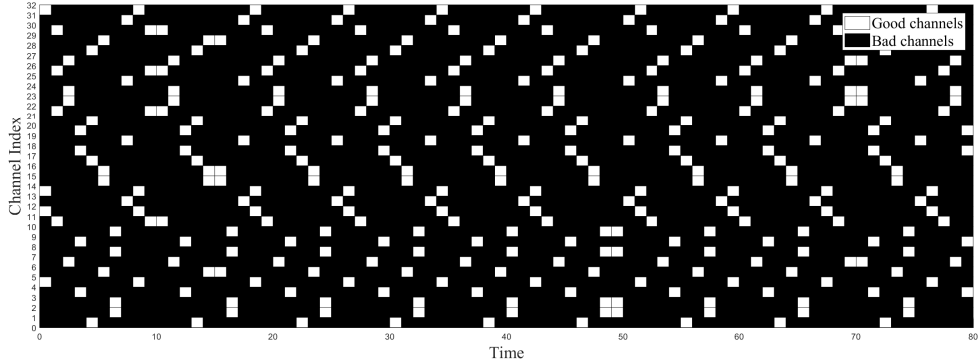}
	\caption{A switching pattern when each four channels of the 32 channels are grouped, with a switching probability $p = 0.9$}
	\label{fig:AbitraryPattern_32_group}
\end{figure}

\subsubsection{Multiple Good Channels}
Now, we consider the switching pattern of a group channels, and in each state in this pattern, there are multiple channels in good state. For instance, a pattern with four channels in good state at a given time is shown in Fig. \ref{fig:AbitraryPattern_32_group}. In this experiment, we fix the switching probability at $p = 0.9$, and study the performance in terms of the average sum reward. We assume that the user is allowed to access more than one channel at a time, and for each selected channel, the user will receive a reward ($1$ or $-1$). To show how many good channels are selected on average in one iteration, we sum the reward received in each iteration and average over time. In the implementation, we assume there are always $6$ good channels when the total number of channels is $16$, and $12$ good channels when the total number of channel is $32$.

\begin{figure}
	
	\begin{minipage}{.46\linewidth}
		\centering
		\subfloat[16 channels]{\label{fig:multiaction_A}\includegraphics[width=1.15\textwidth]{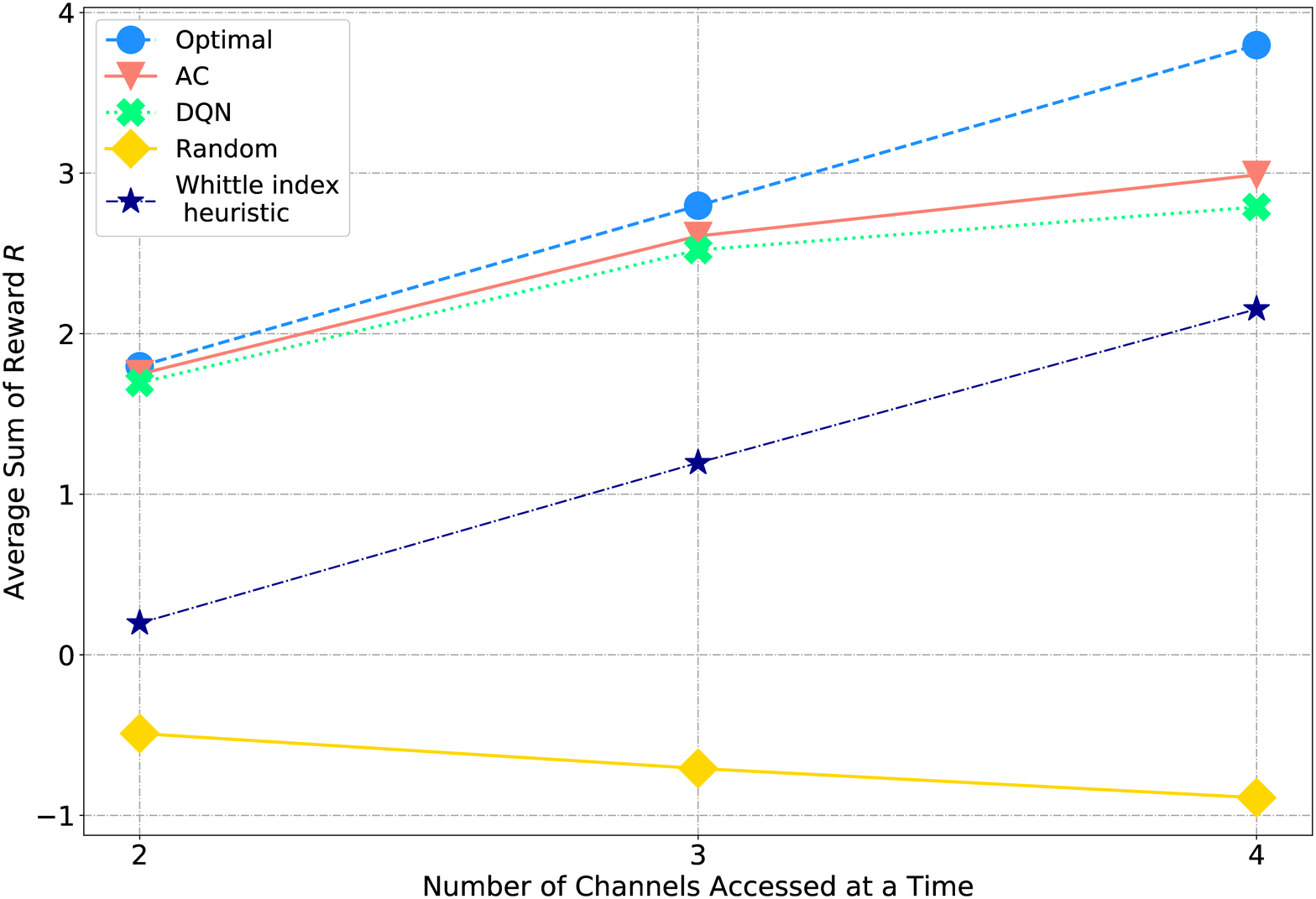}}
	\end{minipage}%
	\begin{minipage}{.46\linewidth}
		\centering
		\subfloat[32 channels]{\label{fig:multiaction_B}\includegraphics[width=1.15\textwidth]{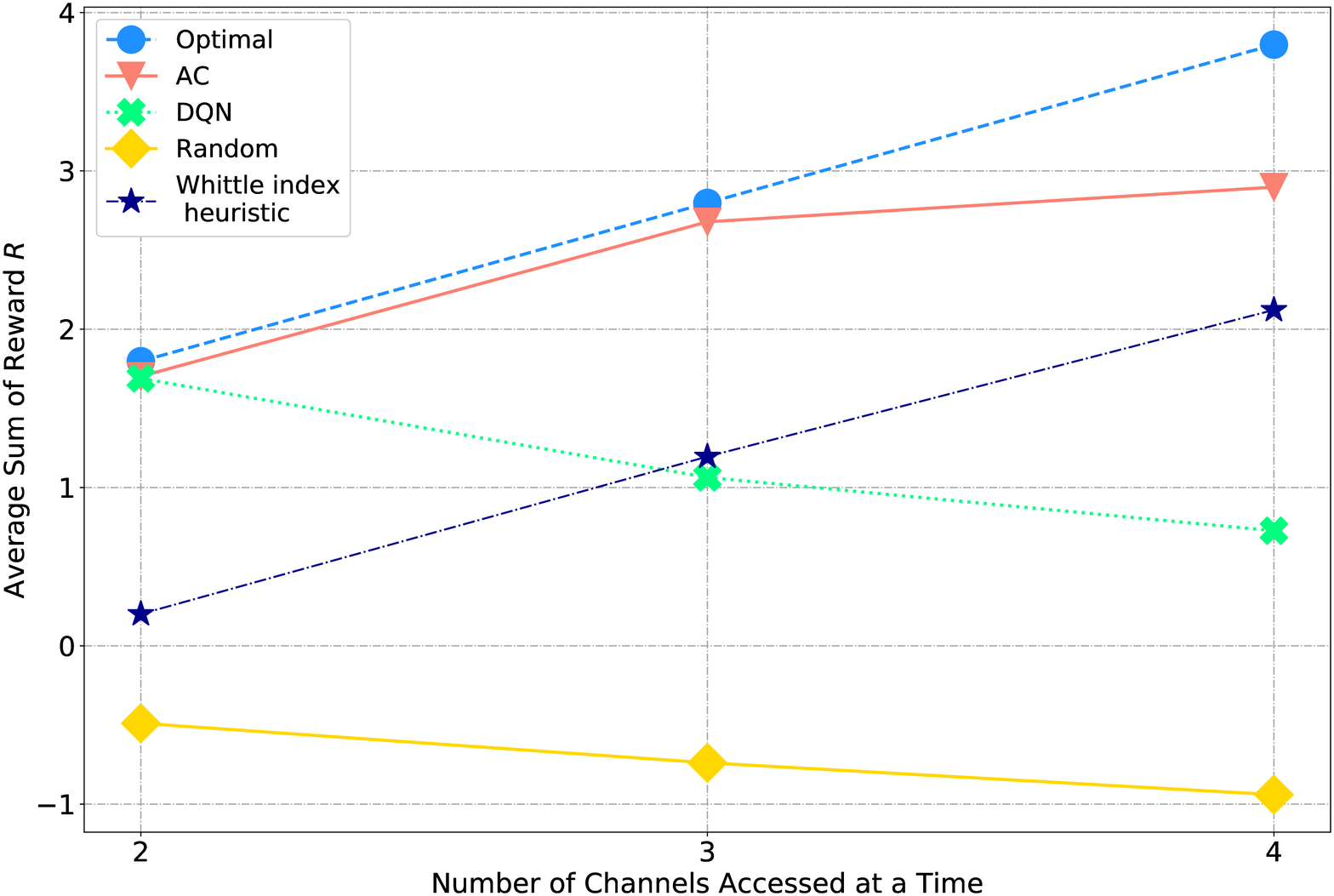}}
	\end{minipage}\par\medskip
	
	\caption{Average sum of reward vs. number of channels that can be accessed at a time.}
	\label{fig:multiaction}
\end{figure}

In Figs. \ref{fig:multiaction}(a) and \ref{fig:multiaction}(b), we plot the performance when the user can access $2$, $3$, and $4$ different channels at a time among $16$ and $32$ channels, respectively. The performances of the optimal policy with known channel dynamics is always around $0.9$ times the maximum reward for all scenarios because of the value of the switching probability. We observe that random access overall performs poorly due to not learning the switching patterns. Whittle index heuristic achieves higher average sum rewards. Interestingly, when the number of channels that can be accessed at a time increases, the performance of Whittle index heuristic exceeds that of DQN in the experiment with $32$ channels. Among the learning-based policies, the actor-critic agent achieves highest rewards. For both actor-critic and DQN policies, when there are $16$ channels, the average sum reward increases as the number of channels that can be accessed increases but with diminishing returns. As introduced in Section \ref{sec: system model}, when the user is allowed to choose $k$ different channels, each action stands for a set of channels to be accessed. Therefore, the size of the action space grows from $N$ to $N \choose k$. And the performance of the learning based policies is significantly influenced by the size of the action space. For instance, in the case with $32$ channels, the average sum reward achieved by the DQN agent diminishes as the number of channels to be accessed increases, demonstrating that the DQN agent is not able to handle the growing size of the action space. On the other hand, the average sum reward received by the actor-critic agent is still slightly increasing, showing the capability of the actor-critic reinforcement learning algorithm in working with relatively large action spaces.


\subsection{Average Reward in the Multi-User Case}
In this subsection, we provide simulation results for the multi-user scenario. As introduced in Algorithm \ref{alg:AC-MU}, we propose a decentralized multi-agent framework to solve the problem, which allows each user to make its own decision. In this experiment, each user can only access one channel and is unaware of other users' decisions, meaning that there could be collisions.
So, to maximize the reward, each agent is required to learn not only the channel switching pattern, but also the other users' channel access patterns to avoid collisions.

\subsubsection{Multi-User Scenario without Priorities}
\begin{figure}
	\centering
	\includegraphics[width=0.8\linewidth]{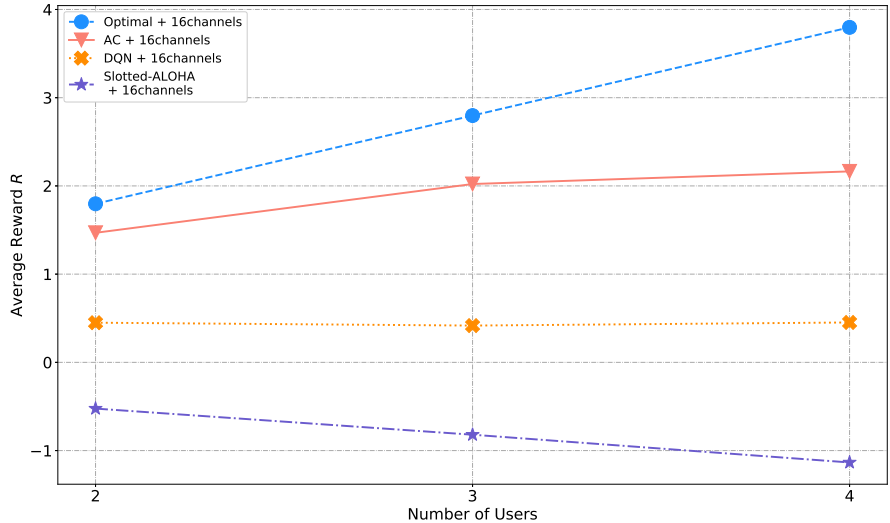}
	\caption{Average reward vs. number of users}
	\label{fig:decentralized}
\end{figure}

First, we consider a scenario in which there are $m$ users, where $m = {2,3,4}$, and no priority is assigned to any user. We run the proposed actor-critic agent and the DQN agent, assuming that there are $16$ channels with $6$ good channels in each state within the switching pattern, and the switching probability is fixed at $p = 0.9$. As a reference, we also evaluate the performance of the optimal policy with known channel switching patterns, and the slotted-ALOHA where each user employs the random access policy independently. We again consider the average sum reward as the performance metric. As shown in Fig. \ref{fig:decentralized}, the optimal policy ensures that users avoid choosing the same channel in each time slot, and hence the optimal average sum reward is actually the same as that achieved in the case of a single user accessing multiple different channels. The averaged reward received by slotted-ALOHA keeps decreasing as the number of users increases, which means that in the slotted-ALOHA policy, the collisions cannot be effectively avoided. For the decentralized actor-critic multi-agent policy, the tendency of the performance is very similar to that of the AC curve shown in Fig. \ref{fig:multiaction}(a), but the values of the average sum reward are smaller, due to the absence of information on other users. As to the performance of the DQN agent, the average sum reward is rather low and varies only slightly when the number of users increases, indicating that the agent is not capable in this decentralized multi-user channel  selection scenario.


\subsubsection{Multi-User Scenario with Priorities} \label{sec:multi-user-priorities}
Now, we address the multi-user case where there are $3$ users and $16$ channels, and assume that one of these three users has higher priority than the other two. The user with the higher priority is referred to as the primary user, and the other two are secondary users. Again, we assume that there are always enough channels for users to transmit, however, some of the channels are more favorable compared to the others in the sense that they have improved channel conditions and have greater channel capacity. We refer to the channels that can provide better transmission quality as excellent channels, and the other available channels can again be in good or bad states. Hence, we now have an extended model in which the channels can be in one of the three states: \emph{excellent, good,} and \emph{bad}. The decentralized agents are expected to be able to find the good channels and take advantages of the excellent channels.

To encourage the users to access the excellent channels, we assume that the reward for the excellent channels are doubled. To give the priority to the primary users, we assume that the reward received by the primary user will also be doubled regardless of whether the reward is positive or negative. In our experiments, we assume $2$ excellent channels and $4$ good channels in each channel state.

 \paragraph{Primary User Sharing the Channel with Secondary Users}
In this part, we consider the scenario that the primary user will share the channels with the secondary users in the presence of a collision. Here, we assign each user an index, and the user $1$ is chosen as the primary user. Then we record the channel access result of each user. Here, we mark the results using $5$ labels:
\begin{itemize}
	\item Excellent Channels: The user selects an excellent channel and occupies it alone.
	\item Collision in Excellent Channels: Two or three users access the same excellent channel.
	\item Good Channels: The user selects a good channel and occupies it alone.
	\item Collision in Good Channels: Two or three users access the same good channel.
	\item Bad Channels: The user selects a bad channel.
\end{itemize}

\begin{figure}
\centering
\includegraphics[width=.7\linewidth]{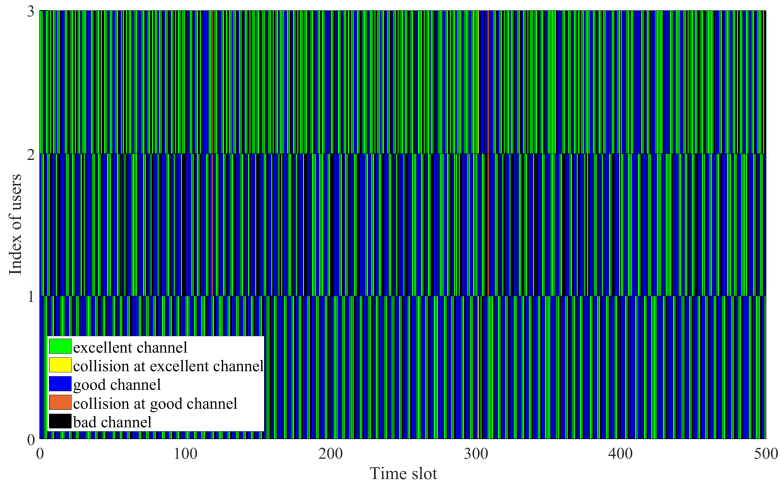}
\caption{The channel selection results based on the \underline{decentralized actor-critic agents} of all users over time in the case that the primary user shares the channel with secondary users in case of a collision.}
\label{fig:ac_primary}
\end{figure}

\begin{figure}
\centering
\includegraphics[width=.7\linewidth]{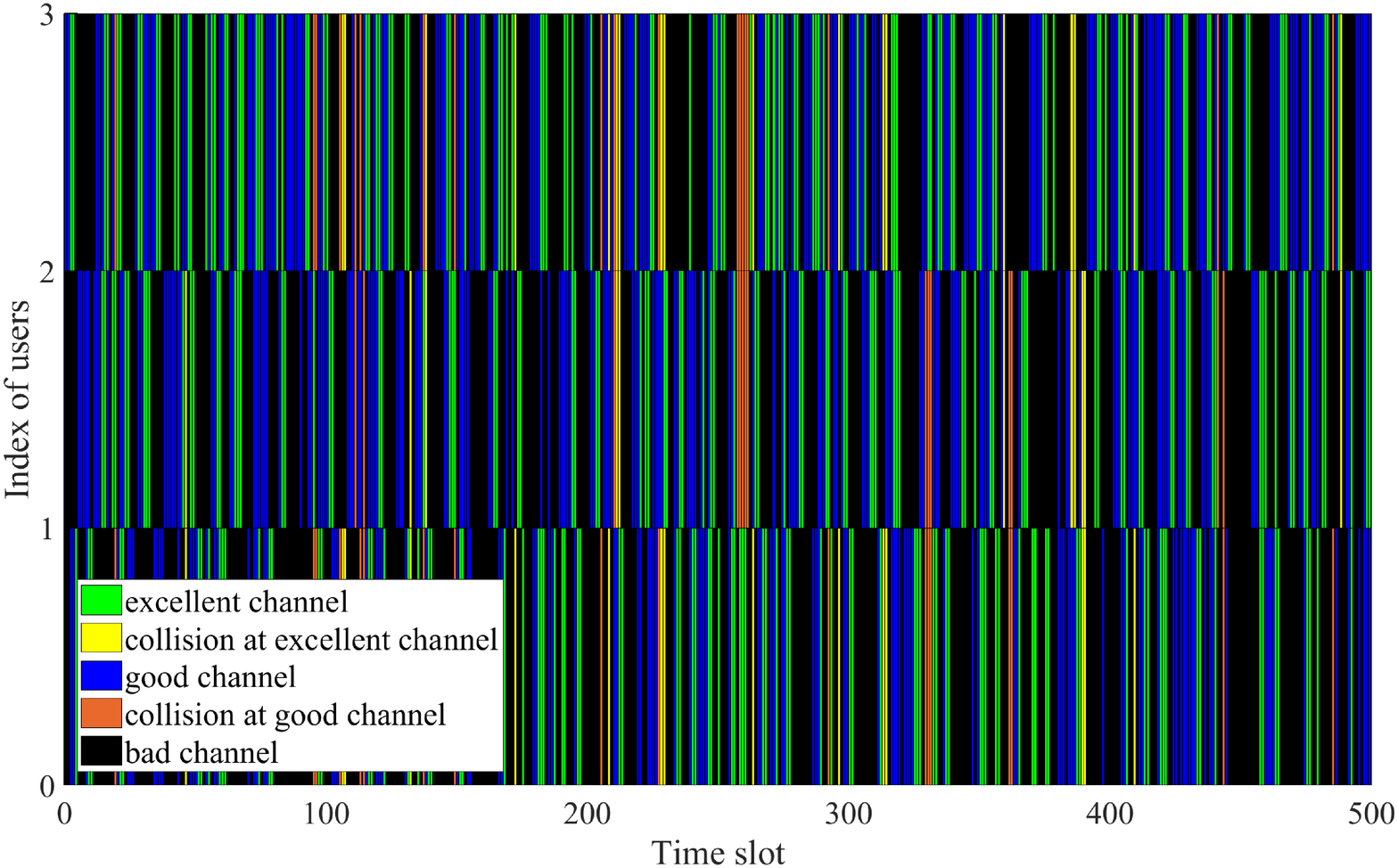}
\caption{The channel selection results based on the \underline{decentralized DQN agents} of all users over time in the case that the primary user shares the channel with secondary users in case of a collision.}
\label{fig:dqn_primary}
\end{figure}

Fig. \ref{fig:ac_primary} and Fig. \ref{fig:dqn_primary} present the users' channel access patterns based on the proposed actor-critic agents and the DQN agents in a period of $500$ time slots, respectively. And we summarize the distribution (or equivalently the computed probabilities) of different channel access results in Table \ref{table:ACP1} and Table \ref{table:dqnP1} for the two channel access frameworks, respectively. It is obvious that the proposed actor-critic is more competitive in selecting excellent channels and good channels. Also the lower probabilities of collisions at excellent and good channels indicate that the proposed framework is effective in learning other users' decision patterns to avoid collisions. As to the users' priorities,  we find that the proposed actor-critic agents do not necessarily guarantee that the primary user occupies the excellent channels most of the  time. One explanation is that even though the reward of the primary user is doubled, the other users still try to access the excellent channels to achieve their own maximum reward. Another reason is that the negative reward of the primary user is also doubled, and hence there is a chance that the primary user will be less aggressive to avoid such increased penalty. This is evidenced by the observation that the primary user attains the minimum probability of experiencing a bad channel.

\begin{table*}[]
	\centering
	\caption{The distribution of different channel access results for \underline{decentralized actor-critic agents} of all users over time in the case that the primary user shares the channel with secondary users in case of a collision.}
	\label{table:ACP1}
	\resizebox{\textwidth}{!}{
		\begin{tabular}{|c|c|c|c|c|c|}
			\hline
			User Index & Excellent Channels & Collision at Excellent Channels & Good Channels & Collision at Good Channels & Bad Channels \\ \hline
			1          & 0.4240             & 0.0020                          & 0.4720        & 0                          & 0.1020       \\ \hline
			2          & 0.3140             & 0                               & 0.4640        & 0.0040                     & 0.2180       \\ \hline
			3          & 0.5040             & 0.0020                          & 0.2840        & 0.0040                     & 0.2060       \\ \hline
		\end{tabular}
	}
\end{table*}

\begin{table*}[]
	\centering
	\caption{The distribution of different channel access results for \underline{decentralized DQN agents} of all users over time in the case that the primary user shares the channel with secondary users in case of a collision.}
	\label{table:dqnP1}
	\resizebox{\textwidth}{!}{
	\begin{tabular}{|c|c|c|c|c|c|}
		\hline
		User Index & Excellent Channels & Collision at Excellent Channels & Good Channels & Collision at Good Channels & Bad Channels \\ \hline
		1          & 0.1900             & 0.0300                          & 0.2340        & 0.0360                     & 0.5100       \\ \hline
		2          & 0.1900             & 0.0220                          & 0.3240        & 0.0300                     & 0.4340       \\ \hline
		3          & 0.2360             & 0.0360                          & 0.3300        & 0.0380                     & 0.3600       \\ \hline
	\end{tabular}
}
\end{table*}

\begin{figure}
\centering
\includegraphics[width=.7\linewidth]{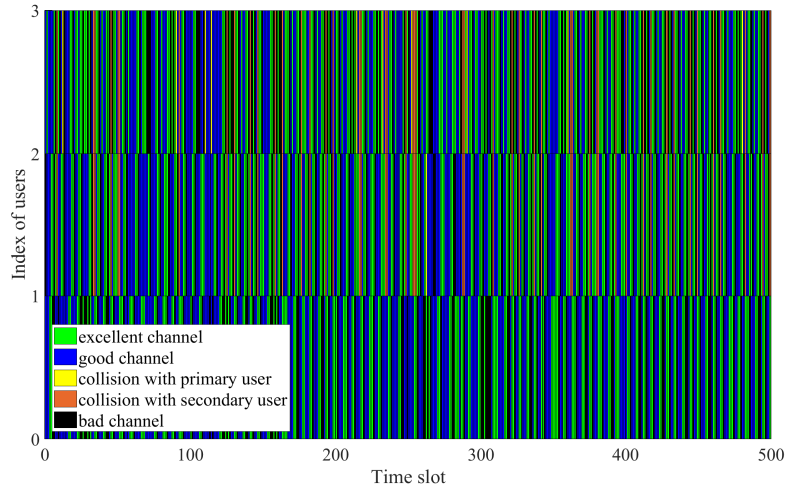}
\caption{Channel selection results based on \underline{decentralized actor-critic agents} of all users over time in the case that the primary user occupies the channel alone in case of a collision.}
\label{fig:ac_primaryDominant}
\end{figure}

\begin{figure}
\centering
\includegraphics[width=.7\linewidth]{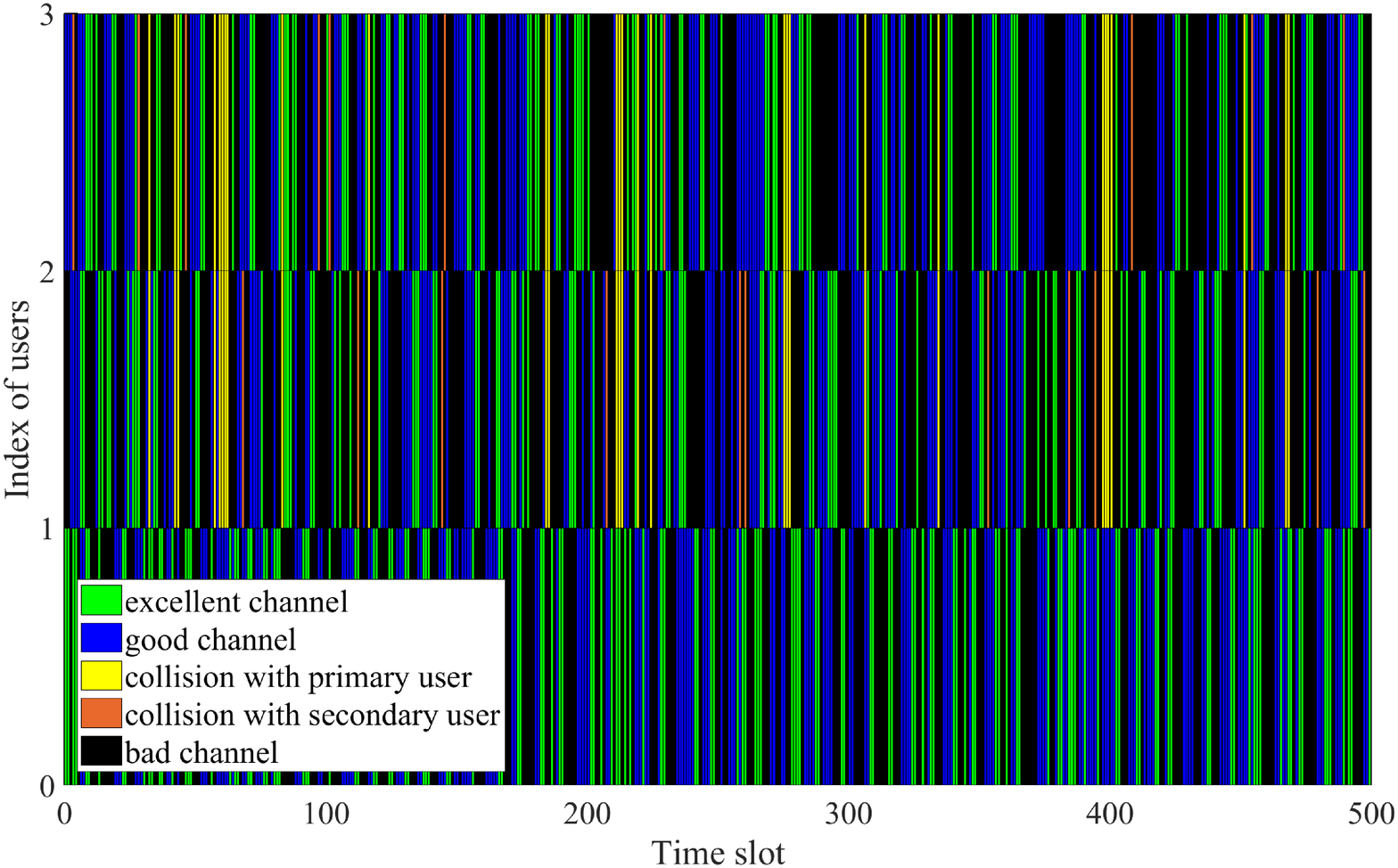}
\caption{Channel selection results based on \underline{decentralized DQN agents} of all users over time in the case that the primary user occupies the channel alone in case of a collision.}
\label{fig:dqn_primaryDominant}
\end{figure}

 \paragraph{Primary User Occupying the Channel Alone in case of a  Collision}
 In this part, we consider the case in which the primary user has the priority to occupy a channel when the secondary users also select it at the same time. Still, user $1$ is assigned to be the primary user, and users $2$ and $3$ are the secondary users. In Figs. \ref{fig:ac_primaryDominant} and \ref{fig:dqn_primaryDominant}, we show the channel access results of all users based on the proposed actor-critic framework and DQN, respectively. And the Table \ref{table:ACP2} and Table \ref{table:DQNP2} summarize the corresponding distribution of results. Since there will not be any collisions occurring from the perspective of the primary user, we have the following cases:
 \begin{itemize}
 	\item Excellent Channels: The user selects an excellent channel and occupies it a alone.
 	\item Good Channels: The user selects a good channel and occupies it alone.
 	\item Collision with the Primary User: The secondary user selects the same excellent/good channel with the primary user.
 	\item Collision with Secondary User: The secondary user selects the same excellent/good channel with the other secondary user.
 	\item Bad Channels: The user selects a bad channel.
 \end{itemize}

With the priority to occupy the channel alone in case of a collision, the probability that the primary user accesses an excellent /good channel is now increased. And from the distribution of the results, we notice that both the actor-critic and DQN polices can effectively enable the secondary user to avoid a collision with the primary user, because in Table \ref{table:ACP2} and Table \ref{table:DQNP2}, the probability of collision with the primary user is much lower than the probability of collision with a secondary user.

\begin{table*}[]
		\centering
		\caption{The distribution of different channel access results for \underline{decentralized actor-critic agents} of all users over time in the case that the primary user occupies the channel alone in case of a collision.}
		\label{table:ACP2}
		\resizebox{\textwidth}{!}{
	\begin{tabular}{|c|c|c|c|c|c|}
		\hline
		User Index & Excellent Channels & Good Channels & \begin{tabular}[c]{@{}c@{}}Collision with \\ Primary User\end{tabular} & \begin{tabular}[c]{@{}c@{}}Collision with \\ Secondary User\end{tabular} & Bad Channels \\ \hline
		1          & 0.4420             & 0.4260        & 0                                                                      & 0                                                                        & 0.1320       \\ \hline
		2          & 0.3780             & 0.3720        & 0.0020                                                                 & 0.1000                                                                   & 0.1480       \\ \hline
		3          & 0.3580             & 0.3180        & 0.0200                                                                 & 0.1000                                                                   & 0.2040       \\ \hline
	\end{tabular}
}
\end{table*}

\begin{table*}[]
		\centering
		\caption{The distribution of different channel access results for \underline{decentralized DQN agents} of all users over time in the case that the primary user occupies the channel alone in case of a collision.}
		\label{table:DQNP2}
		\resizebox{\textwidth}{!}{
	\begin{tabular}{|c|c|c|c|c|c|}
		\hline
		User Index & Excellent Channels & Good Channels & \begin{tabular}[c]{@{}c@{}}Collision with \\ Primary User\end{tabular} & \begin{tabular}[c]{@{}c@{}}Collision with \\ Secondary User\end{tabular} & Bad Channels \\ \hline
		1          & 0.2800             & 0.3520        & 0                                                                      & 0                                                                        & 0.3680       \\ \hline
		2          & 0.1780             & 0.2960        & 0.0220                                                                 & 0.0580                                                                   & 0.4460       \\ \hline
		3          & 0.2000             & 0.2960        & 0.0200                                                                 & 0.0580                                                                   & 0.4260       \\ \hline
	\end{tabular}
}
\end{table*}

\subsection{Time-Varying Environment}
As discussed before, both the proposed actor-critic framework and the DQN framework introduced in \cite{wang2018deep} are reward-driven algorithms which can continually interact with the environment and update the policies. To illustrate the adaptive ability of the proposed framework, we have designed a time-varying environment, where at the beginning, the agent has been trained for pattern $\mathcal{P}_1$, and at time slot $t = 500$, the channel distribution changes to the second pattern $\mathcal{P}_2$, but the change point is unknown to the agent. The experiment was conducted with a fixed switching probability $p = 0.9$, and arbitrary switching order where $32$ channels are grouped into $8$ subsets randomly and each subset has $4$ perfectly correlated channels.

\begin{figure}
	\centering
	\includegraphics[width=.7\linewidth]{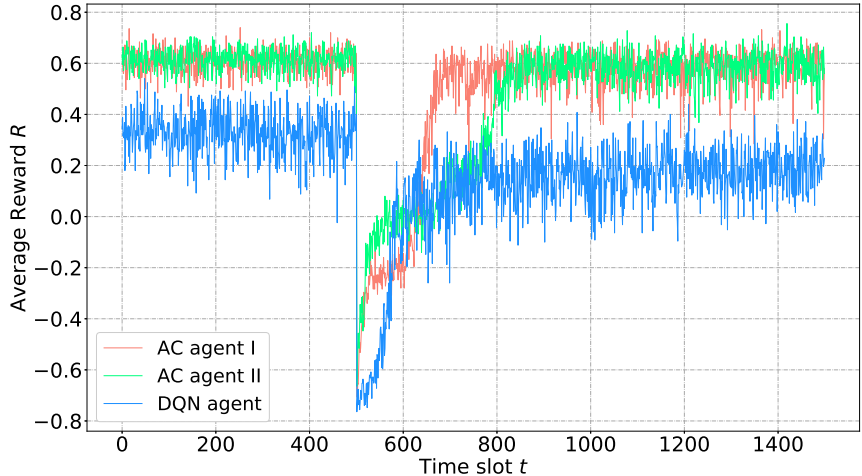}
	\caption{The re-training process in a time-varying environment with the change point at $t = 500$.}
	\label{fig:timeVarying}
\end{figure}

The re-training process is shown in Fig. \ref{fig:timeVarying} in terms of the reward averaged over every $500$ accessing decisions. Considering the learning rate in the actor-critic framework decays as the training process goes by, and the learning rate will influence time needed for the re-training process, we in this experiment test this framework in two different settings: for AC agent I, we allow the agent to reset the learning rate to the initial value when the agent receives negative average reward; and for AC agent II, as a reference, the learning rate will always decay over time. Before the experiment, all agents are well trained and extra time slots are taken to make sure that the learning rates in AC agents are smaller than initial values. Then we set the time when we start the observation as $t = 0$. When the channel state switching pattern changes at $t = 500$, the average reward achieved by both actor-critic framework and DQN drops to negative values suddenly. And then, over time, the policies get updated and adapt to the new pattern, and as a result, the average rewards gradually increase and reach to the previous levels. In the re-training process, due to the difference in learning rate, the AC agent I is quicker to learn the new pattern while the AC agent II is slower but experiences slightly less fluctuations in terms of the average reward. Both AC agents eventually perform as well as before the change point. Comparing the time duration it takes for the agent to get back the previous level and the performance after the re-training process, we conclude that our proposed framework is very competitive in terms of the adaptive ability, though the actor-critic structure which has two separate neural networks takes slightly more time to converge. We also observe that the DQN agent attains an average reward level that is less than before the change  point.

\subsection{Study of Runtime}
To meet the real-time requirements, the channel access decisions must be made quickly. To highlight the efficiency of the actor-critic framework, we have computed the average runtime needed for making one decision and compared it with that needed in the DQN framework. Table \ref{my-label} shows the runtime for one decision needed by the actor-critic (AC) agent and the DQN agent for the case of having a single good channel out of $N$ channels in total, where $N = \{16, 32, 64\} $.

\begin{table}[]
	\centering
	\caption{The runtime needed for each channel access decision}
	\label{my-label}
	\begin{tabular}{|c|c|c|c|l}
		\cline{1-4}
		number of channels & AC agent & DQN agent & \% reduced &  \\ \cline{1-4}
		16                 & 0.002428 & 0.025381  & 90.4328    &  \\ \cline{1-4}
		32                 & 0.003998 & 0.030833  & 87.0340    &  \\ \cline{1-4}
		64                 & 0.004002 & 0.059308  & 93.2527    &  \\ \cline{1-4}
	\end{tabular}
\end{table}

The proposed actor-critic framework is actually more complicated in architecture because it has two neural networks and hence has more parameters to update. But we only pass one actor to the critic, so that the critic requires less computational resources. Another important reason why our framework can have significant savings in the runtime is that we do not need to replay any experience because the LSTD of the critic network is enough to ensure that the actor policy is updating in the correct direction, while the DQN proposed in \cite{wang2018deep} replayed $32$ samples for each time of updating to make the channel access policy stable. For the current number of channels and users, the second reason for the substantial improvements in runtime is that the action space is limited. But once that action space increases, as the number of channels increases, the first reason will become more significant.

Indeed, to demonstrate the impact of memory replay, we briefly discuss the computational complexities of AC and DQN agents next. For the actor-critic network, let us assume that the number of neurons in each layer $i$ of actor network is $a_i$, and the number of neurons in each layer $j$ of critic network is $c_j$, and there are $A$ layers in actor network and $C$ layers in critic network. We further assume that the input size is $K$. In each iteration, the number of calculations at neurons is $(Ka_1 + \sum_{i = 1}^{A-1} a_i a_{i+1}) + 2\cdot(Kc_1 + \sum_{j = 1}^{C-1} c_j c_{j+1})$.
	
	For the DQN, we assume that the number of neurons in each layer $g$ if $d_g$, and there are $D$ layers in total. Also, we suppose that the minibatch size is $M$. With the same input size $K$, the number of calculations of DQN is $M(Kd_1 + \sum_{g = 1}^{D-1} d_g d_{g+1})$.
	
	If we assume that the actor network has the same size as the DQN, and the critic network has the same size except for the output layer (the size of critic output layer is fixed to be $1$, and the size of actor network and DQN output layer are fixed to be the number of actions), then the ratio of computational complexity between actor-critic network and DQN is approximately $3/M$, where the typical values of $M$ are $16$, $32$, $64$, and for some cases it can be even greater. Therefore, we conclude that not replaying the minibatch is a important reason that can explain the high time efficiency of actor-critic. Also, when the DQN replays the minibatch sample, the time consumption for importing the data is also nonnegligible.

\section{Conclusion}\label{sec : conclusion}
In this work, we have considered the dynamic multichannel access problem modeled as a POMDP. To effectively find the channel access policy, we have proposed and implemented model-free actor-critic deep reinforcement learning frameworks in single-user and multi-user scenarios. We have tested the single agent framework on round-robin and arbitrary switching scenarios, and compared the average reward with that of the DQN framework, random access police, Whittle index heuristic and optimal policy. Also, we have studied the performance of the proposed framework in cases in which multiple different channels are selected simultaneously. We have demonstrated the proposed framework's superior ability in handling a large number of channels, high tolerance against uncertainty, and large action spaces. In the multi-user case, we have addressed models with users operating with or without priorities. For users without priority, we have presented results on the average sum reward to demonstrate the decentralized actor-critic agents' capability to learn the channel switching patterns as well as the other users' action patterns. In the case of users with priority, we have computed the distribution of different channel access results under different channel allocation policies and shown that the proposed framework is competitive in various scenarios. To highlight the adaptive ability, we have conducted simulations in a time-varying environment and demonstrated that the proposed framework learns the new patterns effectively in a relatively short period of time. Finally, we have demonstrated the efficiency of the actor-critic framework by computing the percentage of runtime that can be saved compared to the DQN framework.

\end{spacing}



\ifCLASSOPTIONcaptionsoff
  \newpage
\fi

\begin{spacing}{1.2}

\bibliographystyle{ieeetr}
\bibliography{Channel_selection}

%
%
%
%
%
%
\end{spacing}

\end{document}